%% file: main.tex
\documentclass[lettersize,journal]{IEEEtran}
\usepackage{amsmath,amsfonts}
\usepackage{algorithmic}
\usepackage{algorithm}
\usepackage{array}
\usepackage{subfigure}
\usepackage{textcomp}
\usepackage{stfloats}
\usepackage{url}
\usepackage{verbatim}
\usepackage{graphicx}
\usepackage{cite}
\usepackage{tcolorbox}
\usepackage{color}
\usepackage[colorlinks = true]{hyperref}
\definecolor{darkblue}{RGB}{1,96,173}
\definecolor{darkpurple}{RGB}{84,39,134}
\definecolor{darkorange}{RGB}{223,77,2}
\definecolor{darkgreen}{RGB}{12,131,56}

\usepackage[utf8]{inputenc} % allow utf-8 input
\usepackage[T1]{fontenc}    % use 8-bit T1 fonts
\usepackage{hyperref}       % hyperlinks
\usepackage{url}            % simple URL typesetting
\usepackage{booktabs}       % professional-quality tables
\usepackage{amsfonts}       % blackboard math symbols
\usepackage{nicefrac}       % compact symbols for 1/2, etc.
\usepackage{microtype}      % microtypography
\usepackage{xcolor}         % colors
\usepackage{graphicx} %插入图片的宏包
\usepackage{float} %设置图片浮动位置的宏包
\usepackage{makecell}
\usepackage{multirow}
\usepackage{multicol}
\usepackage[figuresright]{rotating}
\usepackage{enumitem}

\usepackage{stackengine}
\usepackage{scalerel}
\usepackage{svg}

\usepackage{fontawesome}

\def\ie{$i.e.$}
\def\eg{$e.g.$}
\def\etc{$etc$}
\newcommand{\wrt}{\textit{w.r.t.~}}
\newcommand{\vs}{\textit{v.s.~}}
\long\def\comment#1{}
\usepackage{xcolor}

\begin{document}

\title{BackdoorBench: A Comprehensive Benchmark and Analysis of Backdoor Learning}

\author{
Baoyuan Wu, Hongrui Chen, Mingda Zhang, Zihao Zhu, Shaokui Wei, Danni Yuan, Mingli Zhu, Ruotong Wang, Li Liu, Chao Shen
\\
\IEEEcompsocitemizethanks{
%\protect\\
%E-mail: \{lilongkang, wubaoyuan\}@cuhk.edu.cn.
\IEEEcompsocthanksitem 
The first seven authors are with School of Data Science, The Chinese University of Hong Kong, Shenzhen, Guangdong, 518172, P.R. China, email: wubaoyuan@cuhk.edu.cn, hongruichen@link.cuhk.edu.cn, mingda zhang@link.cuhk.edu.cn, zihaozhu@link.cuhk.edu.cn, shaokuiwei@link.cuhk.edu.cn, danniyuan@link.cuhk.edu.cn, minglizhu@link.cuhk.edu.cn, ruotongwang1@link.cuhk.edu.cn. 
Li Liu is with the Hong Kong University of Science and Technology (Guangzhou), China, email: avrillliu@hkust-gz.edu.cn.
Chao Shen is with the School of Cyber Science and Engineering, 
Xi’an Jiaotong University, email: chaoshen@xjtu.edu.cn. 
\IEEEcompsocthanksitem 
Corresponding author: Baoyuan Wu (wubaoyuan@cuhk.edu.cn).
}
}



\maketitle

\input{sections/abstract}
\input{sections/introduction}

\input{sections/related_work}
\input{sections/our_benchmark}
\input{sections/evaluation}
\input{sections/analysis}

\input{sections/conclusion}

{\small
\bibliographystyle{plain}
\bibliography{backdoor-short-name}
}

% \newpage

% \section*{Biography Section}


% \begin{IEEEbiographynophoto}{Baoyuan Wu}

\vfill

\end{document}

%% file: sections/abstract.tex
\begin{abstract} 
  As an emerging approach to explore the vulnerability of deep neural networks (DNNs), 
  backdoor learning has attracted increasing interest in recent years, and many seminal backdoor attack and defense algorithms are being developed successively or concurrently, in the status of a rapid arms race. 
  However, mainly due to the diverse settings, and the difficulties of implementation and reproducibility of existing works, there is a lack of a unified and standardized benchmark of backdoor learning, causing unfair comparisons or unreliable conclusions (\eg, misleading, biased or even false conclusions). 
  Consequently, it is difficult to evaluate the current progress and design the future development roadmap of this literature. 
  To alleviate this dilemma, we build a comprehensive benchmark of backdoor learning called \textit{BackdoorBench}. 
  Our benchmark makes three valuable contributions to the research community. \textbf{1)} We provide an integrated implementation of representative backdoor learning algorithms (currently including 20 attack and 32 defense algorithms), based on an extensible modular-based codebase. 
  \textbf{2)} We conduct comprehensive evaluations with 5 poisoning ratios, based on 4 models and 4 datasets, leading to 11,492 pairs of attack-against-defense evaluations in total. 
  % \textbf{2)} We conduct comprehensive evaluations of 12 attacks against 16 defenses, with 5 poisoning ratios, based on 4 models and 4 datasets, thus 11,492 pairs of evaluations in total. 
  \textbf{3)} Based on above evaluations, we present abundant analysis from 10 perspectives via 18 useful analysis tools, and provide several inspiring insights about backdoor learning. 
  We hope that our efforts could build a solid foundation of backdoor learning to facilitate researchers to investigate existing algorithms, develop more innovative algorithms, and explore the intrinsic mechanism of backdoor learning. Finally, we have created a user-friendly website at \url{http://backdoorbench.com}, which collects all important information of BackdoorBench, including the link to Codebase, Docs, Leaderboard, and Model Zoo.  

  % Backdoor learning is an emerging and vital topic for studying deep neural networks' vulnerability (DNNs). 
  % Many pioneering backdoor attack and defense methods are being proposed, successively or concurrently, in the status of a rapid arms race. 
  % However, we find that the evaluations of new methods are often unthorough to verify their claims and accurate performance, mainly due to the rapid development, diverse settings, and the difficulties of implementation and reproducibility.  
  % Without thorough evaluations and comparisons, it is not easy to track the current progress and design the future development roadmap of the literature. 
  % To alleviate this dilemma, we build a comprehensive benchmark of backdoor learning called \textit{BackdoorBench}. 
  % It consists of an extensible modular-based codebase (currently including implementations of 8 state-of-the-art (SOTA) attacks and 9 SOTA defense algorithms) and a standardized protocol of complete backdoor learning. 
  % We also provide comprehensive evaluations of every pair of 8 attacks against 9 defenses, with 5 poisoning ratios, based on 5 models and 4 datasets, thus 8,000 pairs of evaluations in total. 
  % We present abundant analysis from different perspectives about these 8,000 evaluations, studying the effects of different factors in backdoor learning. 
  % All codes and evaluations of BackdoorBench are publicly available at \url{https://backdoorbench.github.io}.
\end{abstract}


%% file: sections/introduction.tex
\section{Introduction}
\label{sec: introduction}

% \IEEEPARstart{T}{his} \red{need to update}
\IEEEPARstart{W}{ith}
the widespread application of deep neural networks (DNNs) in many mission-critical scenarios, the security issues of DNNs have attracted more attentions. One of the typical security issue is backdoor learning, which could insert an imperceptible backdoor into the model through maliciously manipulating the training data or controlling the training process.  
It brings in severe threat to the widely adopted paradigm that people often download a unverified dataset/checkpoint to train/fine-tune their models, or even outsource the training process to the third-party training platform.

Although backdoor learning is a young topic in the machine learning community, its development speed is remarkable and has shown the state of a rapid arms race. 
When a new backdoor attack or defense method is developed based on an assumption or observation, it will be quickly defeated or evaded by more advanced adaptive defense or attack methods which break previous assumptions or observations. 
However, we find that the evaluations of new methods are often insufficient, with comparisons with limited previous methods, based on limited models and datasets. 
The possible reasons include the rapid development of new methods, diverse settings (\eg, different threat models), as well as the difficulties of implementing or reproducing previous methods. 
Without thorough evaluations and fair comparisons, it is difficult to verify the real performance of a new method, as well as the correctness or generalization of the assumption or observation it is built upon. 
Consequently, we cannot well measure the actual progress of backdoor learning by simply tracking new methods. 
This dilemma may not only postpone the development of more advanced methods, but also preclude the exploration of the intrinsic reason/property of backdoor learning.

To alleviate this dilemma, we build a comprehensive benchmark of backdoor learning, called \textbf{BackdoorBench}. It is built on an extensible modular based codebase, consisting of the attack module, the defense module, as well as the evaluation and analysis module. Until now, we have implemented 20 representative backdoor attack methods and 32 representative defense methods,  and provided 18 analysis tools (\eg, t-SNE, Shapley value, Grad-CAM, frequency saliency map and neuron activation). More methods and tools are continuously updated. 
To ensure fair and reproducible evaluations, we provide a standardized protocol for the complete backdoor learning procedure, covering data preparation, attack, defense, and output evaluation. Our comprehensive evaluations explores various backdoor methods across 5 poisoning ratios, 4 DNN models, and 4 datasets, leading to  11,492 pairs of attack-against-defense
evaluations in total. BackdoorBench aims to facilitate the design of new methods and the exploration of intrinsic properties in backdoor learning, promoting its development.
% Based on the codebase, to ensure fair and reproducible evaluations, we also provide a standardized protocol of the complete procedure of backdoor learning, covering every step of the data preparation, backdoor attack, backdoor defense, as well as the saving, evaluation and analysis of immediate/final outputs.
% Moreover, we conduct a comprehensive experiment of different backdoor attack and defense methods, with  5 poisoning ratios, based on 4 DNN models and 4 datasets. We hope that BackdoorBench could provide useful tools to facilitate not only the design of new attack/defense methods but also the exploration of intrinsic properties and reasons for backdoor learning, such as to promote the development of backdoor learning.
%, which are summarized in Table \ref{table:datasets}. 

% \input{tables/experiment}
% Moreover, we conduct comprehensive evaluations of every pair of attack and defense method (\ie, 8 attacks against 9 defenses), with 5 poisoning ratios, based on 5 DNN models and 4 databases, thus up to 8,000 pairs of evaluations in total. 
% These evaluations allow us to analyze some characteristics of backdoor learning. In this work, we present the analysis from four perspectives, to study the effects of attack/defense methods, poisoning ratios, datasets and model architectures, respectively. 

Building upon the comprehensive benchmark established above, we conducted further analysis and exploration from the following perspectives. %The analytical aspects primarily unfolded along the following parts. 
\textbf{First}, from the perspective of data, we investigate the four properties: sample selection, poisoning ratio, trigger generalization and stealthiness. 
% \textbf{1) Trigger Generalization}: Different setting of same trigger is applied to discover the ability of generalization between training and inference stage when trigger pattern is different in these two stages. \textbf{2) The stealthiness of backdoor attack}: SSIM between origianl samples and backdoored version with FID between poison samples and target-class samples are utilized to measure the stealthiness from both pixel space and feature space. \textbf{3) Poisoning ratio}: We analysis the different performance behaviour under different poisoning ratio for selected attack-defense pairs, comparing different defense methods behaviour to conclude a trend for different poisoning ratio. \textbf{4)Effect of backdoor sample selection} With different attack methods and different poisoning ratios, we analysis the overall performance of different sample selection methods. 
\textbf{Second}, in terms of model, we investigate how architecture difference leads to different properties of backdoored model and closely examine the loss landscape and hyper-parameter for all attack and defense methods from a perspective of the interaction between attack and defense methods.
% \textbf{5) Model Structure}: We utilized the proportion of each neuron activated by the clean/backdoor sample, to explain the influence of architecture from a feature extraction view. 
\textbf{Third}, we examine the sensitivity of backdoor attack and defense methods to hyper-parameters in order to investigate the nuances of various approaches.
\textbf{Last}, in examining training dynamics, we pay attention to the intriguing phenomena of backdoor attacks, and rapid learning, and employ memory-related analytical tools to delve into the underlying principles of these occurrences. 
% Fourth, considering training dynamics, we focus on the intriguing phenomenon of backdoor attack, quick learning, and use memory-related analytical tools to explore the principles behind phenomena.
% \textbf{6) Analysis of Forgetting}: The number of forgetting event during training is recorded to shade light on the learning dynamics during the training process. \textbf{7) Quick Learning} We provided a detailed analysis by investigating gradient in early stage from a perspective of gradient information. 
% Fourth, , we 
% \textbf{8) Sharpness}: We utilized the hessian matrix information to analysis the landscape and how attack, defense, and poisoning ratio affect landscape sharpness. \textbf{9,10) Analysis of Each Individual Attack and Defense Method}: After extensive comparative analysis, we have also changed the hyperparameters of individual attack/defense methods, delving into a comprehensive exploration of their effectiveness on critical aspects.

This work is extended based on our previous conference paper \cite{backdoorbench}, compared to which the most significant updates and contributions are summarized from the following three aspects:
\begin{itemize}
    \item \textit{Manuscript}: We reorganized the overall structure, and rewrote most important contents. For example, we rewrote Abstract and Introduction, added several latest related works in Section \ref{sec: reated work}, reorganized Tables \ref{tab: 20 attacks} \& \ref{tab: 31 defense algorithms} and added two metrics in Section \ref{sec: benchmark}, and expanded the original Section IV to two new sections (\ie, Sections \ref{sec: evaluation} and \ref{sec: analysis}) by adding much more evaluations and analyses. 
    \item \textit{GitHub repository}: We added implementations of 12 backdoor attack and 23 backdoor defense algorithms, leading to 20 attacks and 32 defenses in total, as well as 13 analysis tools, leading to 18 tools in total. Besides, we organized the codes of all implemented algorithms into classes, updated the data processing code to handle larger-scale datasets more efficiently, and added code of generating poisoned datasets. 
    \item \textit{Website}: We improved the website by updating the leaderboard to cover all new evaluations, and adding a model zoo where all models in our evaluations can be downloaded, as well as a documentation to illustrate the definition and usage of each important function in the GitHub repository for user convenience. 
\end{itemize}
We believe that all above updates make substantial new contributions to the community of backdoor learning, with not only a more comprehensive and user-friendly benchmark and toolbox, but also deep understandings about the intrinsic characteristics and underlying mechanism of backdoor learning.

%% file: sections/related_work.tex
\section{Related Work}
\label{sec: reated work}

% \red{need to update}

\paragraph{Backdoor attacks}
%Since the first proposed work BadNets \cite{gu2019badnets}, several seminal backdoor attack methods have been proposed. 
According to the threat model, existing backdoor attack methods can be partitioned into two general categories, including \textbf{data poisoning} and \textbf{training controllable}. \textbf{1)} \textbf{Data poisoning attack} means that the attacker can only manipulate the training data. Existing methods of this category focuses on designing different kinds of triggers to improve the imperceptibility and attack effect, including visible (\eg, BadNets  \cite{gu2019badnets}) \vs invisible (\ie, Blended \cite{chen2017targeted}, Refool \cite{liu2020reflection}, LSB \cite{li2020invisible}, WPDA \cite{song2024wpda}) triggers, local (\eg, label consistent attack \cite{turner2019labelconsistent,zhao2020clean}) vs global (\eg, SIG \cite{SIG} Color backdoor \cite{jiang2023color}) triggers, additive (\eg, Blended \cite{chen2017targeted}) \vs non-additive triggers (\eg, 
% smooth low frequency (LF) trigger \cite{zeng2021rethinking}, 
FaceHack \cite{sarkar2022facehack}, frequency-based adaptive trigger \cite{yu2023backdoor}), sample agnostic (\eg, BadNets \cite{gu2019badnets}) \vs sample specific (\eg, SSBA \cite{ssba}, Sleeper agent \cite{souri2021sleeper}, Poison Ink \cite{zhang2022poison}) triggers, \etc. The definitions of these triggers can be found in the bottom notes of Table \ref{tab: 20 attacks}. In addition, the pattern of label modification of poisoned samples is also worth paying attention to, including label inconsistent attacks (\eg, BadNets all-to-all \cite{gu2019badnets}, Marksman \cite{doan2022marksman}, FLIP \cite{jha2023flip}) \vs label consistent attacks (label consistent attack(LC)\cite{turner2019labelconsistent} Sleeper agent\cite{souri2021sleeper}).
\textbf{2)} \textbf{Training controllable attack} means that the attacker can control both the training process and training data simultaneously. Consequently, the attacker can learn the trigger and the model weights jointly, such as LIRA \cite{doan2021lira}, Blind  \cite{bagdasaryan2021blind}, WB \cite{doan2021backdoor}, Input-aware \cite{nguyen2020input}, WaNet \cite{nguyen2021wanet}, BppAttack \cite{wang2022bppattack}, IBA \cite{nguyen2023iba} \etc. 
% We refer the readers to some backdoor surveys \cite{gao2020backdoor,liu2020survey} for more backdoor attack methods. 
Besides, there is another one special setting of training-controllable backdoor attacks, \ie, backdoor attack against federated learning \cite{konevcny2016federated}.
In that setting, the attacker can only control the local training data and influence the local models at some malicious clients, such as DBA \cite{xie2019dba}, Neurotoxin \cite{zhang2022neurotoxin}, and \cite{wang2020attack-fed}, \cite{chen2020backdoor}, \cite{fung2020limitations}, \cite{bhagoji2019analyzing},  \cite{bagdasaryan2020backdoor}.

\paragraph{Backdoor defences} According to the defence stage in the life-cycle of machine learning system, existing defence methods can be partitioned into three categories, including \textbf{pre-training-stage}, \textbf{training-stage}, \textbf{post-training-stage} and \textbf{inference-stage} defenses. 
\textbf{1)} \textbf{Pre-training-stage defense} means that the defender aims to prevent backdoor injection before the training stage. One natural choice is identifying poisoned samples from the untrustworthy training dataset, also called \textit{poisoned sample detection} (PSD). The key point of PSD is finding or designing some distriminative metrics to distinguish poisoned samples from clean samples. To achieve that, existing PSD methods have attempted to utilize activation distribution (\eg, AC \cite{ac}, SPECTRE \cite{spectre}, SCAn \cite{demon}, Beatrix \cite{beatrix}), activation path (\eg, NC \cite{wang2019neural}), activation gradient (\eg, AGPD \cite{yuan2023activation}), activation sensitivity to input transformation (\eg, FCT \cite{chen2022dbr}), training loss (\eg, ABL \cite{li2021anti}, ASSET \cite{asset}), prediction stability to model variation (\eg, CT \cite{qitowards}, and \cite{du2020robust}), and visual-linguistic inconsistency (\eg, VDC \cite{zhu2023vdc}), \etc. 
\textbf{2)} \textbf{Training-stage defense} aims to train a clean model on a untrustworthy training dataset, while without backdoor injection. Existing methods of this type often replace the standard end-to-end supervised training by some well-designed multi-step training procedures, which generally conduct poisoned sample detection and model weight learning simultaneously or alternatively, such as ABL \cite{li2021anti}, DBD \cite{huang2022backdoor}, D-ST \cite{chen2022dbr}, CBD \cite{zhang2023backdoor}, NAB \cite{liu2023beating}, \etc. 
\textbf{3)} \textbf{Post-training-stage defense} describes that given a untrustworthy model, the defender aims to detect whether this model contains backdoor or not (\ie, backdoor detection), or remove/mitigate backdoor from the model (\ie, backdoor removal/mitigation). Here we only review the latter type. There are two general approaches in existing methods, including structure modification and tuning-based approach. The structure modification approach could prune backdoor-related neurons (\eg, FP \cite{fp}, ANP \cite{wu2021adversarial}, ShapPruning \cite{guan2022few}, Purifier \cite{zhang2022purifier}, CLP \cite{clp-2022}), or augment some neurons to neutralize the effect of backdoor-related neurons (\eg, NPD \cite{zhu2023neural}). 
The tuning-based approach often adjusts the model weights through some well-designed fine-tuning (\eg, NC \cite{wang2019neural}, i-BAU \cite{i-bau}, FT-SAM \cite{zhu2023enhancing}, FST \cite{min2023towards}, SAU \cite{wei2023shared}, NAD \cite{nad-iclr-2020}).
\textbf{4)} \textbf{Inference-stage defense} aims to prevent the backdoor activation during inference, even though the model may contain backdoor. 
To achieve that goal, the defender could choose to detect and reject the poisoned query, \ie, poisoned query/sample detection (\eg, STRIP \cite{gao2019strip}, SentiNet \cite{sentinet}, TeCo \cite{liu2023detecting}, SCALE-UP \cite{guo2023scale}, FreqDetector \cite{zeng2021rethinking}, FREAK \cite{freak}), or give correct prediction corresponding to the clean features rather than the backdoor trigger, \ie, poisoned query/sample recovering (\eg, Orion \cite{orion}, NAB \cite{liu2023beating}, ZIP \cite{shi2023black}).

We would like to refer the readers to two latest surveys \cite{wu2024attacks,wu2023defenses} about backdoor attacks and defenses for more details, respectively.  

\input{tables/table-attack-algorithms}

\paragraph{Related benchmarks} 
Several libraries or benchmarks have been proposed for evaluating the adversarial robustness of DNNs, such as CleverHans \cite{papernot2016cleverhans}, Foolbox  \cite{rauber2017foolbox,rauber2017foolboxnative}, AdvBox \cite{goodman2020advbox}, RobustBench \cite{croce2020robustbench}, RobustART \cite{tang2021robustart}, ARES \cite{dong2020benchmarking}, Adversarial Robustness Toolbox (ART) \cite{ART-IBM}, \etc. 
However, these benchmarks mainly focused on adversarial examples \cite{fgsm,kurakin2018adversarial}, which occur in the testing stage. 
In contrast, there are only a few libraries or benchmarks for backdoor learning (\eg,  TrojAI \cite{karra2020trojai}, TrojanZoo \cite{trojanzoo2022} and openbackdoor \cite{cui2022unified}).
%, mainly due to that backdoor learning is relatively new compared to adversarial examples, and backdoor learning involves more terms (\eg, data, model and training). 
%Specifically, two existing benchmarks are related to our BackdoorBench, including TrojAI \cite{karra2020trojai} and TrojanZoo \cite{trojanzoo2022}.
Specifically, the most similar benchmark is TrojanZoo, which implemented 8 backdoor attack methods and 14 backdoor defense methods. However, there are significant differences between TrojanZoo and our BackdoorBench in two main aspects. \textbf{1) Codebase}: Compared with TrojanZoo, which implements many inheritance classes, we use a primarily flat structure and avoid using many inheritance classes. This flat structure makes our code easier to understand, especially for academic purposes. By doing this, we have ensured that our code is user-friendly and does not require a heavy learning load. This approach prioritizes simplicity and makes it easier for people in academic settings to grasp and use our code effectively and quickly. %although both benchmarks adopt the modular design to ensure easy extensibility, TrojanZoo adopts the object-oriented programming (OOP) style, where each module is defined as one class. In contrast, BackdoorBench adopts the procedural oriented programming (POP) style, where each module is defined as one function, and each specific algorithm is implemented by several functions in a streamline. %Although both programming styles have their own advantages, the learning cost for beginners of BackdoorBench is lower than that of TrojanZoo, and the implementation of new methods based on BackdoorBench is easier than that based on TrojanZoo.   
\textbf{2) Analysis and findings}: TrojanZoo has provided very abundant and diverse analysis of backdoor learning, mainly including the attack effects of trigger size, trigger transparency, data complexity, backdoor transferability to downstream tasks, and the defense effects of the trade-off between robustness and utility, the trade-off between detection accuracy and recovery capability, the impact of a trigger definition. In contrast, BackdoorBench provides 10 new analyses from 4 perspectives that cover main components of backdoor learning, as well as 18 analysis tools. % mainly including the effects of poisoning ratios and number of classes, the quick learning of backdoor, trigger generalization, memorization and forgetting of poisoned samples, as well as several analysis tools. %Due to the space limit, more details about the differences between two benchmarks will be presented in \textbf{Appendix}. 
In summary, we believe that BackdoorBench could provide new contributions to the backdoor learning community, and the competition among different benchmarks is beneficial to the development of this topic.

%% file: tables/table-attack-algorithms.tex
\begin{table*}[htbp]
%\vspace{-1em}
\centering
\caption{Categorizations of 20 implemented backdoor attack algorithms in BackdoorBench, according to threat model, trigger and target label.}
\label{tab: 20 attacks}
\scalebox{0.84}{
\begin{tabular}{p{.095\textwidth}
% >{\centering\arraybackslash}  
p{.05\textwidth}
% >{\centering\arraybackslash} 
p{.05\textwidth}
% >{\centering\arraybackslash}  
p{.04\textwidth}
% >{\centering\arraybackslash} 
p{.04\textwidth}
% >{\centering\arraybackslash}  
p{.04\textwidth}
% >{\centering\arraybackslash} 
p{.04\textwidth} % >{\centering\arraybackslash}
p{.04\textwidth} %>{\centering\arraybackslash} 
p{.05\textwidth} %>{\centering\arraybackslash}  
p{.05\textwidth} %>{\centering\arraybackslash} 
p{.04\textwidth} %>{\centering\arraybackslash}  
p{.038\textwidth} %>{\centering\arraybackslash} 
p{.034\textwidth} %>{\centering\arraybackslash} 
p{.034\textwidth} %>{\centering\arraybackslash} 
p{.038\textwidth} %>{\centering\arraybackslash} 
p{.036\textwidth} %>{\centering\arraybackslash}
}
\hline
 \multirow{2}{*}{\textbf{Attack}}  & \multicolumn{2}{c}{\textbf{Threat model}} & \multicolumn{2}{c}{\textbf{Trigger visibility}} & \multicolumn{2}{c}{\textbf{Trigger coverage}} & \multicolumn{2}{c}{\textbf{Trigger fusion mode I}} & \multicolumn{2}{c}{\textbf{Trigger fusion mode II}} &  
 \multicolumn{2}{c}{\textbf{Trigger fusion mode III}} & \multicolumn{2}{c}{\textbf{Target type}} 
 \\
 \cmidrule(lr){2-3} \cmidrule(lr){4-5} \cmidrule(lr){6-7} \cmidrule(lr){8-9} \cmidrule(lr){10-11} \cmidrule(lr){12-13} \cmidrule(lr){14-15}
 algorithm & \scalebox{0.8}{data} \scalebox{0.8}{poisoning} & \scalebox{0.8}{training}  \scalebox{0.8}{controllable} & visible & invisible & local & global  & additive & non-additive & sample-agnostic & sample-specific & static & dynamic &  \scalebox{0.8}{label} \scalebox{0.8}{inconsistent} & \scalebox{0.8}{label} \scalebox{0.8}{consistent} 
\\
\hline 
BadNets \cite{gu2019badnets} & \checkmark &   & \checkmark &   & \checkmark &  & \checkmark &  & \checkmark & & \checkmark &  & \checkmark &  
\\
Blended \cite{chen2017targeted} & \checkmark &   & & \checkmark   & & \checkmark  & \checkmark &  & \checkmark & & \checkmark &  & \checkmark & 
\\
LC \cite{turner2019labelconsistent} & \checkmark &   & \checkmark &   & \checkmark &  & \checkmark &  & \checkmark & & \checkmark & & & \checkmark 
\\
SIG \cite{SIG} & \checkmark &   & & \checkmark   & & \checkmark  & \checkmark &  & \checkmark &  & \checkmark &  &  & \checkmark
\\     
LF \cite{zeng2021rethinking} & \checkmark &   & & \checkmark  &  & \checkmark  &  \checkmark &  & \checkmark &   & \checkmark &  & \checkmark &  
\\
SSBA \cite{ssba} & \checkmark &   & & \checkmark   & & \checkmark  &  & \checkmark &  &  \checkmark &  & \checkmark & \checkmark & 
\\
Blind \cite{bagdasaryan2021blind} &  & \checkmark & \checkmark &  & \checkmark &  & \checkmark &  & \checkmark &  & \checkmark &  & \checkmark & 
\\
BppAttack \cite{wang2022bppattack} & & \checkmark &  & \checkmark &  & \checkmark &  & \checkmark &  & \checkmark & \checkmark &  & \checkmark & 
\\
TrojanNN \cite{liu2018trojaning} & & \checkmark &  \checkmark &  &  \checkmark &  & \checkmark &  & \checkmark &  &  \checkmark &  & \checkmark &     
\\
LIRA \cite{doan2021lira} & & \checkmark &  & \checkmark &  & \checkmark &  & \checkmark &  & \checkmark & \checkmark &  & \checkmark & 
\\
\scalebox{0.85}{Input-aware} \cite{nguyen2020input} & &  \checkmark   & \checkmark &   & \checkmark &  & \checkmark &  & & \checkmark &  & \checkmark & \checkmark & 
\\
WaNet \cite{nguyen2021wanet} & &  \checkmark &  & \checkmark &  & \checkmark & & \checkmark & & \checkmark & \checkmark &  & \checkmark & 
\\
CTRL \cite{li2023embarrassingly} & \checkmark &   &  & \checkmark  &  & \checkmark &  & \checkmark & \checkmark & & \checkmark &  &  &  \checkmark
\\
FTrojan \cite{wang2022invisible} & \checkmark &   &  & \checkmark  &  & \checkmark &  & \checkmark & \checkmark & & \checkmark &  & \checkmark &  
\\
ReFool \cite{liu2020reflection} & \checkmark &   &  & \checkmark  &  & \checkmark & \checkmark &  & \checkmark & &  & \checkmark &  &  \checkmark
\\
\scalebox{0.85}{PoisonInk} \cite{zhang2022poison} &  & \checkmark  &  & \checkmark  &  & \checkmark &  & \checkmark &  & \checkmark &  & \checkmark & \checkmark &  
\\
\scalebox{0.8}{Adap-Blend} \cite{adapt} & \checkmark &   & & \checkmark   & & \checkmark  & \checkmark &  & \checkmark & &  & \checkmark & \checkmark & 
\\
TaCT \cite{demon} & \checkmark& & \checkmark& & \checkmark& & \checkmark& & \checkmark & & \checkmark& & \checkmark& 
\\
BELT \cite{qiu2023belt} & \checkmark& & \checkmark & & \checkmark & & \checkmark& & \checkmark& & & \checkmark& \checkmark&
\\
Marksman \cite{doan2022marksman} & \checkmark& \checkmark&  \checkmark & &  & \checkmark & \checkmark & & \checkmark& &  & \checkmark & & \checkmark
\\
\hline
\end{tabular}
}
\end{table*}

%% file: sections/our_benchmark.tex
\section{Our benchmark}
\label{sec: benchmark}

\input{sections/our_benchmark/methods}
\input{sections/our_benchmark/codebase}

\input{sections/our_benchmark/website}

%% file: sections/our_benchmark/methods.tex
\input{tables/table-defense-algorithms}

\subsection{Implemented backdoor learning algorithms}
\vspace{-0.2em}

\noindent 
\paragraph{Selection criterion of implemented algorithms}
Although there have been many backdoor learning algorithms, we hold two basic criterion to select implemented algorithms. 
\textbf{1) Representative algorithms}: It should be a classic (\eg, BadNets) or advanced method (\ie, published in recent top-tier conferences/journals in machine learning or security community). The classic method serves as the baseline, while the advanced method represents the state-of-the-art, and their difference could measure the progress of this field. 
\textbf{2) Easy implementation and reproducibility}: The algorithm should be easily implemented and reproducible. We find that some existing algorithms involve several steps, and some steps depend on a third-party algorithm or a heuristic strategy. Consequently, there are too many hyper-parameters, or randomness of the produced results even with the same setting of hyper-parameters, causing the difficulty on implementation and reproduction. Such methods are not included in BackdoorBench.

% As shown in Table \ref{tab: 8 attacks}, the eight implemented backdoor attack methods cover two mainstream threat models, and with diverse triggers. Among them, BadNets\cite{gu2019badnets}, Blended\cite{chen2017targeted} and LC\cite{turner2019labelconsistent} (label consistent attack) are three classic attack methods, while the remaining 5 are recently published methods. The general idea of each method will be presented in the \textbf{Supplementary Material}. 

% The basic characteristics of 9 implemented backdoor defense methods are summarized in Table \ref{tab: 8 defense}, covering different inputs and outputs, different happening stages, different defense strategies. The motivation/assumption/observation behind each defense method is also briefly described in the last column. More detailed descriptions will be presented in the \textbf{Supplementary Material}. 

\noindent 
\paragraph{Descriptions of backdoor attack algorithms}
According to the taxonomy presented in a recent survey about the attack in adversarial machine learning \cite{wu2024attacks}, we categorize the 20 implemented backdoor attack algorithms from four perspectives, \ie, the threat model, trigger type, trigger fusion mode and target type. As shown in Table \ref{tab: 20 attacks}, these 20 algorithms are diverse enough to cover different types of backdoor attacks. The brief illustration of each individual algorithm is presented in the \textbf{Supplementary Material}. 

\noindent 
\paragraph{Descriptions of  backdoor defense algorithms}
% \label{appendix: sec descriptions of  backdoor defense algorithms}

According to the taxonomy presented in a recent survey about the defense in adversarial
machine learning \cite{wu2023defenses}, we categorize the 31 implemented
backdoor defense algorithms from four perspectives, \ie, the defense stage, input, output, and defense strategy. As shown
in Table \ref{tab: 31 defense algorithms}, these 32 algorithms are diverse enough to cover
different types of backdoor defenses. A brief illustration of
each individual algorithm is presented in the \textbf{Supplementary Material}. %below.

%% file: tables/table-defense-algorithms.tex
\begin{table*}[htbp]
%\vspace{-1.5em}
\centering
\caption{Categorizations of 32 implemented backdoor defense algorithms in BackdoorBench, according to four perspectives, including \textit{defense stage}, \textit{input}, \textit{output},  and \textit{defense strategy}.}
\label{tab: 31 defense algorithms}
\scalebox{0.87}{
\begin{tabular}{
p{.08\textwidth} %>{\centering\arraybackslash} 
p{.03\textwidth} %>{\centering\arraybackslash}  
p{.03\textwidth} %>{\centering\arraybackslash} 
p{.03\textwidth} %>{\centering\arraybackslash} 
p{.03\textwidth} %>{\centering\arraybackslash} 
p{.05\textwidth} %>{\centering\arraybackslash}  
p{.054\textwidth} %>{\centering\arraybackslash}  
p{.05\textwidth} %>{\centering\arraybackslash}   
p{.04\textwidth} %>{\centering\arraybackslash}  
p{.04\textwidth} %>{\centering\arraybackslash}   
p{.05\textwidth} %>{\centering\arraybackslash}  
p{.065\textwidth} %>{\centering\arraybackslash}  
p{.05\textwidth} %>{\centering\arraybackslash}   
p{.065\textwidth} %>{\centering\arraybackslash}  
p{.055\textwidth} %>{\centering\arraybackslash}  
p{.055\textwidth} %>{\centering\arraybackslash}  
}
\hline
 \multirow{2}{*}{\textbf{Defense}} & \multicolumn{4}{c}{\textbf{Defense stage}}  & \multicolumn{3}{c}{\textbf{Input}} & \multicolumn{2}{c}{\textbf{Output}} & \multicolumn{6}{c}{\textbf{Defense strategy}} 
\\
\cmidrule(lr){2-5} \cmidrule(lr){6-8} \cmidrule(lr){9-10} \cmidrule(lr){11-16}   
algorithm &\scalebox{0.9}{pre-} &\scalebox{0.95}{in-} & \scalebox{0.86}{post-} & \scalebox{0.72}{inference} & \scalebox{0.85}{backdoored} & \scalebox{0.85}{subset of} & \scalebox{0.85}{poisoned} & \scalebox{0.95}{secure} & \scalebox{0.95}{benign} & backdoor & backdoor & poison & trigger & backdoor & backdoor  
\\
  & training & training & training & & \scalebox{0.85}{model} & \scalebox{0.85}{benign dataset} & dataset & model & dataset & detection & identification & detection & identification & mitigation & inhibition 
\\
 \hline 
FT & &  & \checkmark & & \checkmark & \checkmark &  & \checkmark &  &  &  &  &  & \checkmark & 
\\
% \hline
FP \cite{fp} & &  & \checkmark & & & \checkmark & \checkmark & \checkmark &  &  & \checkmark &  &  & \checkmark & 
\\
% \hline
NAD \cite{nad-iclr-2020} & &  & \checkmark & & \checkmark & \checkmark  &  & & \checkmark &  &  &  &  & \checkmark & 
\\
% \hline
NC \cite{wang2019neural} & &  & \checkmark & & \checkmark & \checkmark &  & & \checkmark & \checkmark &  &  & \checkmark & \checkmark & 
\\
% \hline
ANP \cite{wu2021adversarial} & & & \checkmark & & \checkmark & \checkmark &  & \checkmark &  &  & \checkmark &  &  & \checkmark & 
\\

AC \cite{ac} & & & \checkmark &  & & \checkmark & & \checkmark & \checkmark &  &  & \checkmark &  & \checkmark & 
\\
% \hline
SS \cite{tran2018spectral} & & & \checkmark & & & & \checkmark  & \checkmark & \checkmark &  &  & \checkmark &  & \checkmark & 
\\
% \hline
ABL \cite{li2021anti} & & \checkmark & & & & & \checkmark  & \checkmark & \checkmark &  &  & \checkmark &  & \checkmark & 
\\
% \hline
DBD \cite{huang2022backdoor} & & \checkmark & & & & & \checkmark & \checkmark & \checkmark &  &  & \checkmark &  &  & \checkmark
\\
I-BAU \cite{i-bau} & & & \checkmark & & \checkmark & \checkmark & & \checkmark &  &  &  &  &  & \checkmark &  
\\
CLP \cite{clp-2022} &  & & \checkmark & & \checkmark &  &  &  \checkmark &  & & \checkmark &  &  & \checkmark & 
\\
D-BR \cite{chen2022dbr} &  &  & \checkmark &  & \checkmark &  & \checkmark & \checkmark & \checkmark &  &  &  \checkmark &  & \checkmark &  
\\
D-ST \cite{chen2022dbr} &  & \checkmark &  & &  &  & \checkmark & \checkmark & \checkmark &  &  & \checkmark &  &  &  \checkmark 
\\
EP \cite{ep-bnp-2022} &  &  & \checkmark & & \checkmark &  & \checkmark & \checkmark &  &  & \checkmark &  &  & \checkmark & 
\\
BNP \cite{ep-bnp-2022} &  &  & \checkmark & & \checkmark & \checkmark &  & \checkmark &  &  &  \checkmark &  &  & \checkmark & 
\\
MCR \cite{mcr} &  & & \checkmark & & \checkmark & \checkmark &  & \checkmark &  &  &  &  &  & \checkmark & 
\\
\scalebox{0.75}{FT-SAM}  \cite{zhu2023enhancing} &  & & \checkmark & & \checkmark  & \checkmark &  & \checkmark &  &  &  &  &  & \checkmark & \\
NAB \cite{liu2023beating} & & & & \checkmark & \checkmark & & & & \checkmark & & & \checkmark & & & \\
NPD \cite{zhu2023neural} &  & & \checkmark & & \checkmark  & \checkmark &  & \checkmark &  &  &  &  &  & \checkmark &  \\
SAU \cite{wei2023shared} &  & & \checkmark & & \checkmark  & \checkmark &  & \checkmark &  &  &  &  &  & \checkmark & \\

RNP \cite{li2023reconstructive} & & & \checkmark & & \checkmark & \checkmark &  & \checkmark &  &  & \checkmark &  &  & \checkmark &\\

SCAn \cite{demon} & \checkmark & & & &  & \checkmark & \checkmark &   &\checkmark & &  & \checkmark & & &\\ 
Beatrix \cite{beatrix}  & \checkmark & & & &  & \checkmark & \checkmark &   &\checkmark & &  & \checkmark & & &\\ 
\scalebox{0.7}{SPECTRE} \cite{hayase2021spectre} & \checkmark & & & &  & \checkmark & \checkmark &   &\checkmark & &  & \checkmark & & \checkmark & \\ 
\scalebox{0.8}{FREAK} \cite{freak} &  & & & \checkmark & \checkmark & \checkmark &  &  & \checkmark &  &  & \checkmark &  &  & \\
AGPD \cite{yuan2023activation} & \checkmark & & & &  & \checkmark & \checkmark &   &\checkmark & &  & \checkmark & & &\\ 
\scalebox{0.7}{STRIP-inference\cite{gao2019strip}}   &  & & & \checkmark & \checkmark & \checkmark & &   & \checkmark & &  & \checkmark & & &\\ 
\scalebox{0.6}{STRIP-pretraining\cite{gao2019strip}}   & \checkmark & & &  & \checkmark & \checkmark & \checkmark &   & \checkmark & &  & \checkmark & & &\\ 
TeCo \cite{liu2023detecting}   &  & & & \checkmark & \checkmark & \checkmark &  &   &\checkmark & &  & \checkmark & & &\\ 
\scalebox{0.8}{SentiNet} \cite{sentinet}   &  & & & \checkmark &\checkmark  & \checkmark &  &   &\checkmark & &  & \checkmark & & &\\ 
AVEA \cite{guo2022aeva} & & & \checkmark &  & \checkmark & \checkmark & & \checkmark & &  \checkmark & & & & & \\
\scalebox{0.6}{SCALE-UP} \cite{guo2023scale}& & &  & \checkmark & \checkmark & & & &\checkmark & & & \checkmark & & &\\
\hline
\end{tabular}
\vspace{-1.5em}
}
\end{table*}

%% file: sections/our_benchmark/codebase.tex
\begin{figure*}[!t] %H为当前位置，!htb为忽略美学标准，htbp为浮动图形
\centering %图片居中
\includegraphics[width=0.95\textwidth]{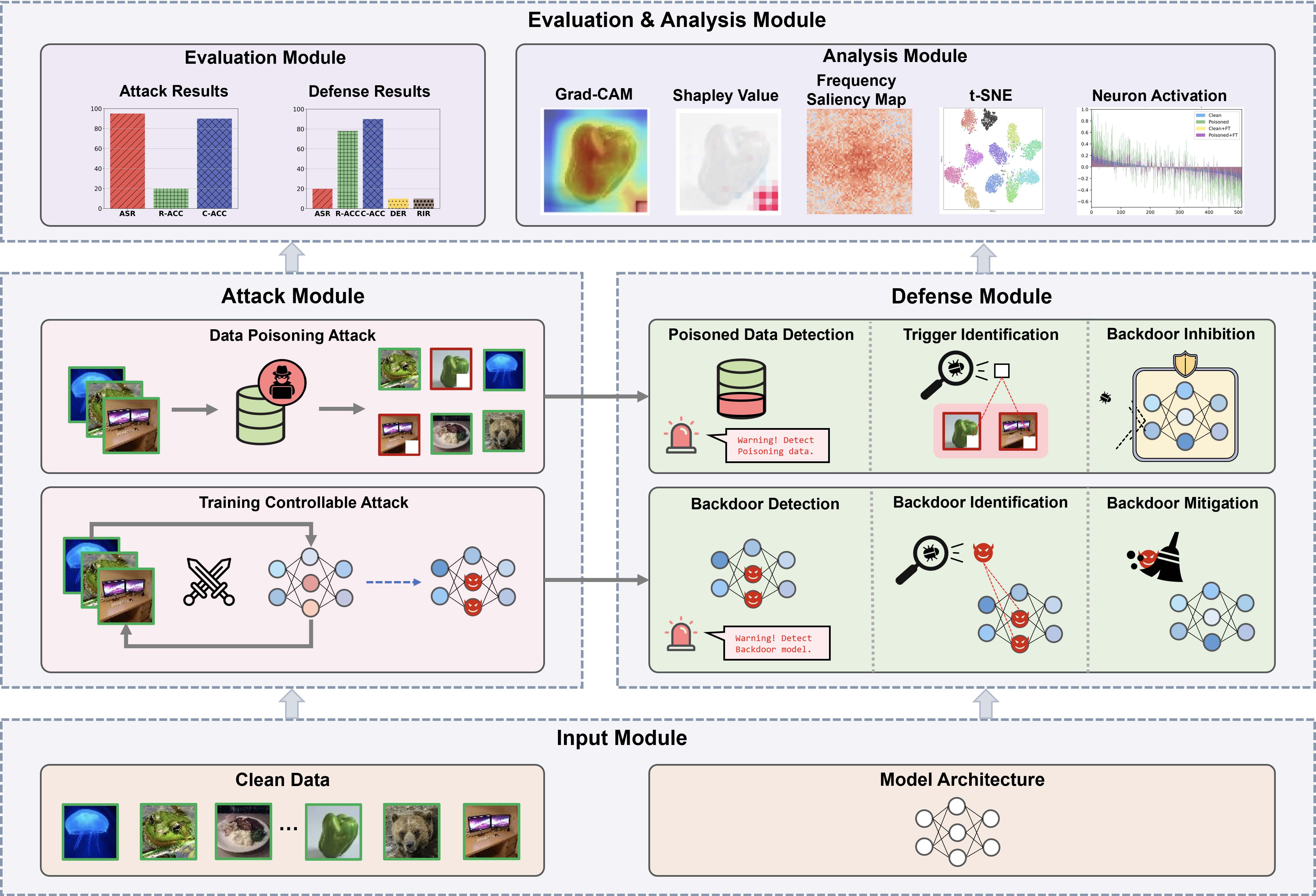} 
% \includesvg[width=1\textwidth]{figs/framework2.svg}
\vspace{-0.9em}
\caption{The general structure of the modular based codebase of BackdoorBench. } 
\label{fig:framework} 
\vspace{-1.5em}
\end{figure*}

\vspace{-0.2em}
\subsection{Codebase}
\label{sec: codebase}
\vspace{-0.2em}

%plot one figure: module-based codebase, with flow to show the protocol  \red{(Mingda)}

We have built an extensible modular-based codebase as the basis of BackdoorBench, which is collected in the Github repository \url{https://github.com/SCLBD/BackdoorBench}.
As shown in Figure \ref{fig:framework}, the main modules contain \textit{input module} (providing clean data and model architectures), \textit{attack module}, \textit{defense module} and \textit{evaluation module} and \textit{analysis module}.

%\textbf{Input module} provides the clean data and model architectures for later attack or defense implementations. 

\textbf{Attack module}. 
%\subsubsection{Attack module}
In the attack module, we provide two sub-modules to implement attacks of two threat models, \ie, \textit{data poisoning} and \textit{training controllable} (see Table \ref{tab: 20 attacks}), respectively. 
For the first sub-module, it provides some functions of manipulating the provided set of clean samples, including trigger generation, poisoned sample generation (\ie, inserting the trigger into the clean sample), and label changing. It outputs a poisoned dataset with both poisoned and clean samples. For trigger generation, we provide a unified protocol; users can simply define a map from image to image to implement a new data poisoning-based trigger. Then, based on label changing and given the poisoning ratio, we can automatically select indexes for poisoned sample generation of the whole dataset without tedious manual operation.
For the second sub-module, given a set of clean samples and a model architecture, it provides two functions of learning the trigger and model parameters, and outputs a backdoored model and the learned trigger. \comment{While our attack framework isn't frozen, it's purposefully crafted for adaptability. This design enables users to effortlessly incorporate our functions and classes without limitations, facilitating smooth customization and extended implementation. }Additionally, we offer a convenient dataset class for manipulating backdoor samples and a fundamental training framework easily adaptable to individual goals. Including these user-friendly tools boosts flexibility and efficiently saves users' time, simplifying the implementation process.

\textbf{Defense module}. 
According to the outputs produced by the attack module, there are also two sub-modules to implement backdoor defenses. 
If given a poisoned dataset, the first sub-module provides three functions of \textit{poisoned sample detection} (\ie, determining whether a sample is poisoned or clean), \textit{trigger identification} (\ie, identifying the location in the poisoned sample), \textit{backdoor inhibition} (\ie, training a secure model through inhibiting the backdoor injection). 
If given a backdoored model, as well as a small subset of clean samples (which is widely required in many defense methods), the second sub-module provides three functions of \textit{backdoor detection} (\ie, determining whether a model has a backdoor or not), \textit{badckdoor identification} (\ie, identifying the neurons in the backdoored model that are related to the backdoor effect), \textit{backdoor mitigation} (\ie, mitigating the backdoor effect from the backdoored model). 
We have updated our defense methods by incorporating modular `subparts', each representing essential functions. These functions serve as the fundamental components, offering a highly adaptable and user-friendly approach. Users can easily select and combine these functions to tailor their defense methods, addressing specific challenges with precision and effectiveness. This detailed approach ensures accuracy and effectiveness in crafting personalized defense methods.
% In extending our defense methods, we've adopted a modular strategy, breaking down our implemented defenses into practical 'subparts.' \comment{This approach offers users a heightened level of adaptability and ease in constructing their personalized defense strategies.} Breaking down defense methods into specific functions lets users choose and combine functions in a detailed and focused manner, resulting in more potent and tailored defense methods. \comment{This flexibility ensures users can efficiently tackle specific challenges in their given contexts, resulting in more potent and tailored defense methods.}

\textbf{Evaluation module}.  
We provide \textbf{5 evaluation metrics}. 
The first three are regular metrics, including \textit{clean accuracy} (\textbf{C-Acc}, \ie, the prediction accuracy of clean samples), \textit{attack success rate} (\textbf{ASR}, \ie, the prediction accuracy of poisoned samples to the target class), and \textit{robust accuracy} (\textbf{R-Acc}, \ie, the prediction accuracy of poisoned samples to the original class).
Note that R-Acc satisfies that ASR + R-Acc $\leq 1$, and lower ASR and higher R-Acc indicate better defense performance. 
In addition, we also provide two new metrics that measures the defense performance more comprehensively. 
One is \textit{defense effectiveness rate} (\textbf{DER}), which was firstly defined in \cite{zhu2023enhancing}. It measures the effectiveness of defense against attack, considering the drops on both ASR and C-Acc, and it is formulated as follows: 
$$ \text{DER} = \frac{\max(0,\Delta_{\text{ASR}}) - \max(0,\Delta_{\text{C-Acc}}) + 1}{2} \in [0,1], $$  
where $\Delta_{\text{ASR}} = \text{ASR}_{bd} - \text{ASR}_{defense}$ represents the drop of ASR after defense, and $\Delta_{\text{C-Acc}} = \text{C-Acc}_{bd} - \text{C-Acc}_{defense}$ denotes the drop of C-Acc. The higher DER value means the better defense performance, corresponding larger ASR drop and smaller C-Acc drop.  
The other is \textit{robust improvement rate} (\textbf{RIR}), which is firstly defined in this work. It measures the recovery performance of defense against attack, considering both the improvement on R-Acc and the drop on C-Acc, and it is formulated as follows:
$$\text{RIR} = \frac{\max(0,-\Delta_{\text{R-Acc}}) - \max(0,\Delta_{\text{C-Acc}}) + 1}{2} \in [0,1], $$ 
where $ \Delta_{\text{R-Acc}} = \text{R-ACC}_{bd} - \text{R-ACC}_{defense}$ indicates the R-ACC drop after defense. 
The higher RIR value implies the better recovery performance, corresponding larger R-ACC improvement and smaller C-Acc drop.  

\textbf{Analysis module}. 
We provide \textbf{18 analysis tool} to facilitate the analysis and understanding of backdoor learning. \textit{t-SNE} and \textit{UMAP} provides a global visualization of feature representations of a set of samples in a model, and it can help us to observe whether the backdoor is formed or not. \textit{Image Quality} evaluates the given results using some image quality metrics.
\textit{Confusion Matrix} gives a deeper breakdown of the model's performance than a single C-Acc value, which helps to find out each class contribution to the overall performance. \textit{Network Structure} provides a detailed visualization of network structure of a given model. \textit{Metrics} gives a comprehensive visualization with all implemented metrics.
\textit{Gradient-weighted class activation mapping (Grad-CAM)} \cite{grad-cam} and \textit{Shapley value map} \cite{NIPS2017_7062} are two individual analysis tools to visualize the contributions of different pixels of one image in a model, and they can show that whether the trigger activates the backdoor or not. 
We also propose the \textit{frequency saliency map} to visualize the contribution of each individual frequency spectrum to the prediction, providing a novel perspective of backdoor from the frequency space. 
% The definition will be presented in \textbf{Supplementary Material}.
\textit{Neuron activation} calculates the average activation of each neuron in a layer for a batch of samples. It can be used to analyze the activation path of poisoned and clean samples, as well as the activation changes \wrt the model weights' changes due to attack or defense, providing deeper insight behind the backdoor. 
\textit{Activated Image} finds the top images who activate the given layer of Neurons most and lists corresponding activation values. It can be beneficial to find out deeper relationship between samples and neurons. It gives a efficient method to understand the contribution of neuron activation from sample perspective.
\textit{Feature Visualization} gives the synthetic images that activate the given layer of Neurons most. All images are generated by Projected Gradient Descend method. This tool gives another perspective to understand neuron activation. \textit{Feature Map} provides the output of a given layer of CNNs for a given image.
\textit{Activation Distribution} gives the class distribution of Top-$k$ images which activate the Neuron most, which analysis from a class-wise perspective.
\textit{Trigger Activation Change} gives the average (absolute) activation change between images with and without triggers for each neuron, which reflects sensitivity of neuron with respect to backdoor features. 
\textit{Lipschitz Constant} gives the lipschitz constant of each neuron. 
\textit{Loss Landscape} gives the visulization of the loss landscape with two random directions.
\textit{Eigenvalues of Hessian} gives the dense plot of hessian matrix for a given batch of data. 
More detailed definition and demo of each analysis tool could be found in our Github repository. 

%\textbf{3 Analysis tools}: We also provide a analysis plugin of three analysis approaches, including T-SNE visualization of the sample representations in the feature space for global analysis, Grad-CAM and Shapley value for individual analysis. \red{In future, we will add more analysis tools for model}

% \textbf{Installations}.  
% We provide two installation approaches, including \textit{source code downloading from github}, and \textit{pip install}. 

\textbf{Protocol}. 
We present a standardized protocol to call aforementioned functional modules to conduct fair and reproducible backdoor learning evaluations, covering every stage from data pre-processing, backdoor attack, backdoor defense, result evaluation and analysis, \etc. 
\comment{We also provide three flexible calling modes, including \textit{pure attack mode} (only calling an attack method), \textit{pure defense mode} (only calling a defense method), as well as \textit{a joint attack and defense mode} (calling an attack against a defense).}
Considering the cognitive burden for users by large-scale projects, we encapsulate various attack and defense methods within inheritable classes, thereby reducing the amount of self-replicating code and enhancing simplicity and ease of adoption.
For attack and defense methods requiring a long workflow, we have systematically decomposed the entire process. Additionally, we have thoughtfully provided users with access to the data and models used in our evaluations for their convenience.
To handle large-scale dataset poisoning, our program supports storing poisoned samples in memory for quick access or locally for constrained memory scenarios.
% Given the potential challenges related to poisoning large-scale datasets, we have incorporated options for data storage locations. When enough memory is available, the entirety of the poisoned samples can be accommodated in memory, facilitating expedited access. Conversely, in instances of high-constrained memory resources, a viable alternative is to save to local storage for the poisoned samples, ensuring our program functionality when faced with extensive datasets.

%% file: sections/our_benchmark/website.tex
\subsection{Website}
\label{sec: website}

We have also created a website at \url{http://backdoorbench.com}, where readers can easily access all useful information of BackdoorBench. It mainly provides \textit{Docs}, \textit{Leaderboard}, \textit{Model zoo}, and the link to Codebase.

\textbf{Docs}. It provides detailed illustrations of the installation, setup of our codebase, the definition of every important function in the codebase, as well as brief demos to illustrate the usage, to help users quickly learn and understand our codebase. 
\comment{In the illustrative tutorial, we comprehensively guide users through crafting a backdoor attack, developing a mitigation defense, and constructing a custom dataset within our framework. Users gain practical insights by covering vital aspects like hyper-parameter configuration, results loading, logger preparation, dataset setup, model initialization, and defense training. We focus on implementing backdoor mitigation techniques in the defense tutorial, and this tutorial also introduces the process of saving defense results for future use. This hands-on guideline equips users with practical knowledge to integrate diverse attacks, defenses, and datasets for specific research objectives.}
The documentation presents our class inheritance hierarchy, outlining the main steps of methods and their sources, along with essential information. 
Moreover, the readers can also find a clear tutorial for introducing how to add new datasets into our benchmark, and build their own backdoor attack and defense algorithms based on our codebase. 
\comment{For example, many data poisoning attack methods adhere to the inheritance structure, from clean training to BadNets and specific methods. On the other hand, defense and detection methods are primarily inherited from the base class of defense. This class plays a crucial role in retrieving necessary information from the attack section, allowing users to develop their methods on top of it.}

\textbf{Leaderboard}. It presents all pairs of attack-against-defense evaluation results we have evaluated. 

\textbf{Model zoo}. We provide a user friendly interface where the reader could choose \textit{Dataset}, \textit{Backbone}, \textit{Attack}, \textit{Defense}, \textit{Poisoning ratio}, such that the corresponding files could be downloaded, including the training and testing poisoned data, 
% training configuration (\ie, settings of all hyper-parameters),
and model checkpoint. With these files and our codebase, users could: 1) reproduce any result reported in the above Leaderboard; 2) investigate more characteristics of the poisoned data and the backdoored model of each attack algorithm, and those of the corresponding ones after each defense algorithm; 3) evaluate their own or any other defense algorithm. 

%% file: sections/evaluation.tex
\section{Evaluation}
\label{sec: evaluation}

\subsection{Evaluation settings}

\textbf{Datasets and models}.  We evaluate our benchmark on 4 widely used dataset, including CIFAR-10 \cite{cifar10_0}, CIFAR-100 \cite{cifar10_0}, GTSRB \cite{gtsrb}, Tiny ImageNet \cite{tinyimagenet}, and 4 model architectures, which is PreAct-ResNet18\cite{preactresnet18}, VGG19-BN\cite{vgg19bn}, ConvNeXt\_tiny\cite{convnext}, ViT\_b\_16\cite{vit}. To fairly measure the performance of all attack and defense methods for each model, no other training tricks are implemented, and we only used the basic training setting for each model. The details of baseline normal training clean accuracy of datasets and models are shown in Table \ref{table:datasets}.

\begin{table*}[ht]
%\vspace{-1em}
\renewcommand\arraystretch{1.5}
\center
\caption{Dataset details and clean accuracy of normal training. }
%\vspace{-1em}
\label{table:datasets}
\resizebox{\linewidth}{!}{
    \begin{tabular}{lccccccc}
    \hline
    \multirow{2}{*}{\textbf{Datasets}} & \multirow{2}{*}{\textbf{Classes}} & \multirow{2}{*}{\textbf{\makecell{Training/\\Testing Size}}} & \multirow{2}{*}{\textbf{Image Size}} & \multicolumn{4}{c}{\textbf{Clean Accuracy}}\\
    \cline{5-8}&&&& PreAct-ResNet18  \cite{preactresnet18}& VGG19-BN \cite{vgg19bn}  & ConvNeXt-tiny\cite{convnext} & ViT-B/16 \cite{vit}\\ 
    \hline
    CIFAR-10 \cite{cifar10_0}                 & 10  & 50,000/10,000 & $32\times 32$ &93.65\% & 92.09\%&92.4\%&96.56\%\\
    CIFAR-100 \cite{cifar10_0}                & 100 & 50,000/10,000 & $64\times 64$ &70.97\%& 66.48\%& 73.46\%&84.59\%\\
    GTSRB   \cite{gtsrb}                  & 43  & 39,209/12,630 & $32\times 32$ & 98.11\%& 98.01\%& 97.60\%&98.84\%\\
    Tiny ImageNet  \cite{Le2015TinyIV}           & 200 & 100,000/10,000& $64\times 64$ &57.81\%& 44.94\%&66.11\%&76.98\%\\ 
    \hline
    \end{tabular}
}
%\vspace{-0.8em}
\end{table*}

\textbf{Attacks and defenses}.  We measure the performance of each pair of 12 attacks against 16 defenses in each setting, and all single attack with defense case. Thus, there are $12 \times 16 = 192$ pairs of evaluations. We consider 5 poison ratios, which is 0.1\%, 0.5\%, 1\%, 5\%, 10\% for each case, based on all 4 datasets and 4 models. Totally we have pairs of evaluations. The performance of every algorithm is measured by the metrics, which are C-Acc, ASR, R-Acc and DER (see Section \ref{sec: benchmark}). The implementation details of all algorithms (\eg, hyper-parameters and computational complexity) are presented in \textbf{Supplementary Material}.

% \textbf{Running environments} Our evaluations are conducted on GPU servers with 2 Intel(R) Xeon(R) Platinum 8170 CPU @ 2.10GHz, RTX3090 GPU (32GB) and 320 GB RAM (2666MHz).
%
% The versions of all involved softwares/packages are clearly described in the README file of the Github repository (see \url{https://github.com/SCLBD/BackdoorBench}). 

\begin{figure*}[ht]
    \centering
    \includegraphics[width=0.95\textwidth]{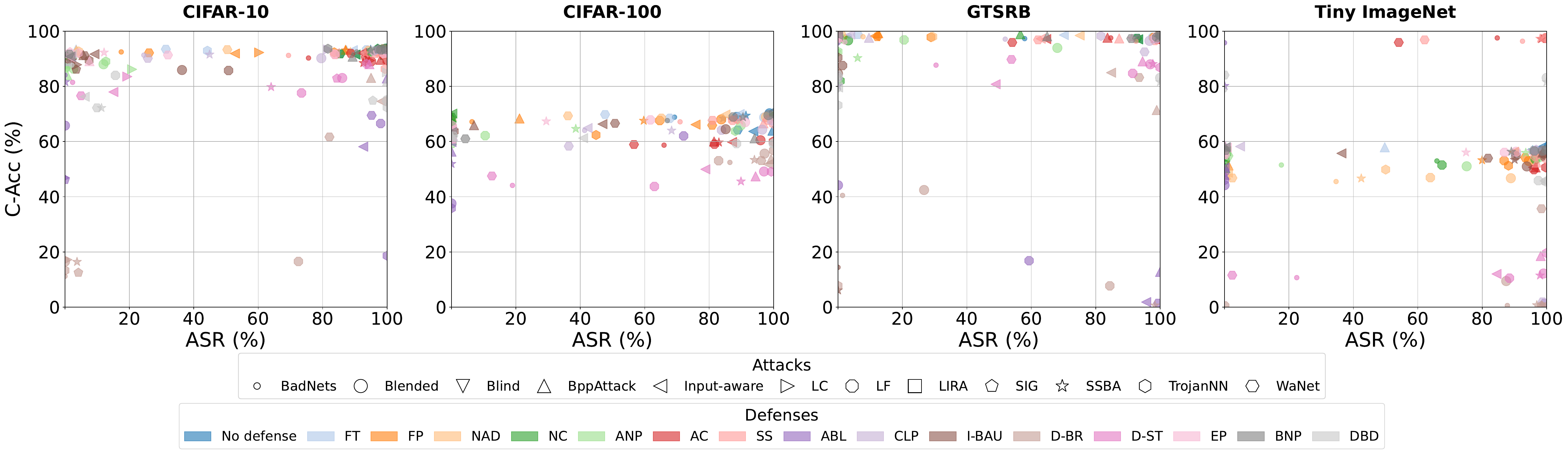}
    \vspace{-0em}
    \caption{Performance distribution of different attack-defense pairs (C-Acc \vs ASR). Each colorful mark represents one attack-defense pair, with attacks distinguished by marks, while defenses by colors.}
    \label{fig:ca-asr}
    \vspace{-0em}
\end{figure*}

\begin{figure*}[ht]
    \centering
    \includegraphics[width=0.95\textwidth]{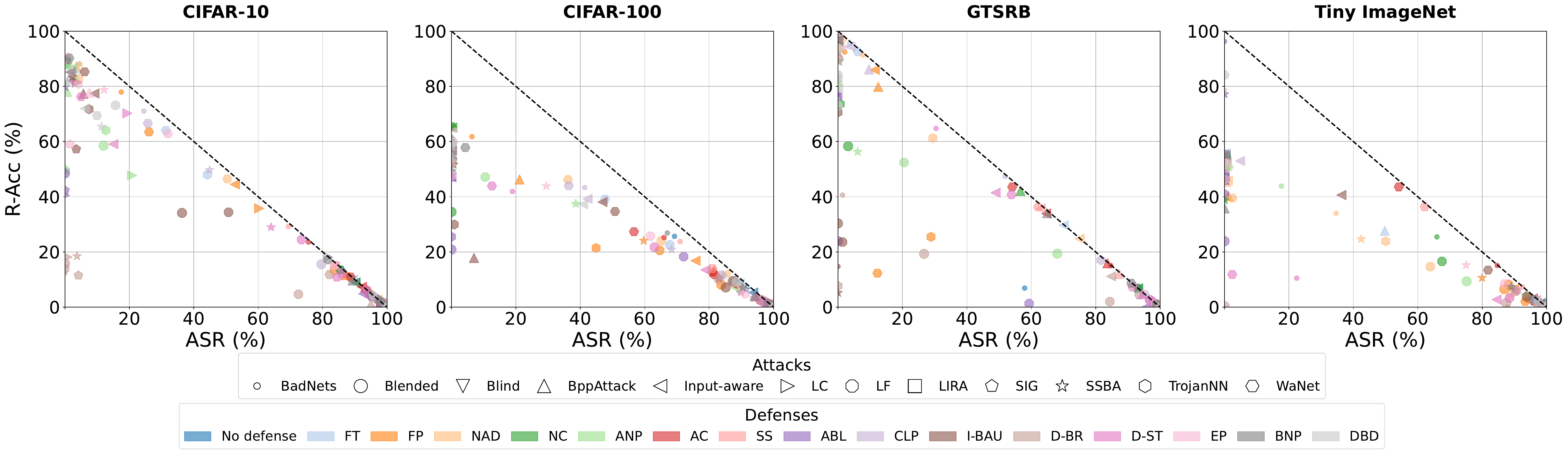}
    \vspace{-1em}
    \caption{Performance distribution of different attack-defense pairs (R-Acc \vs ASR). Each colorful mark represents one attack-defense pair, with attacks distinguished by marks, while defenses by colors.}
    \label{fig:ra-asr}
    \vspace{-1.5em}
\end{figure*}

\begin{figure*}[ht]
    \centering
    \includegraphics[width=0.95\textwidth]{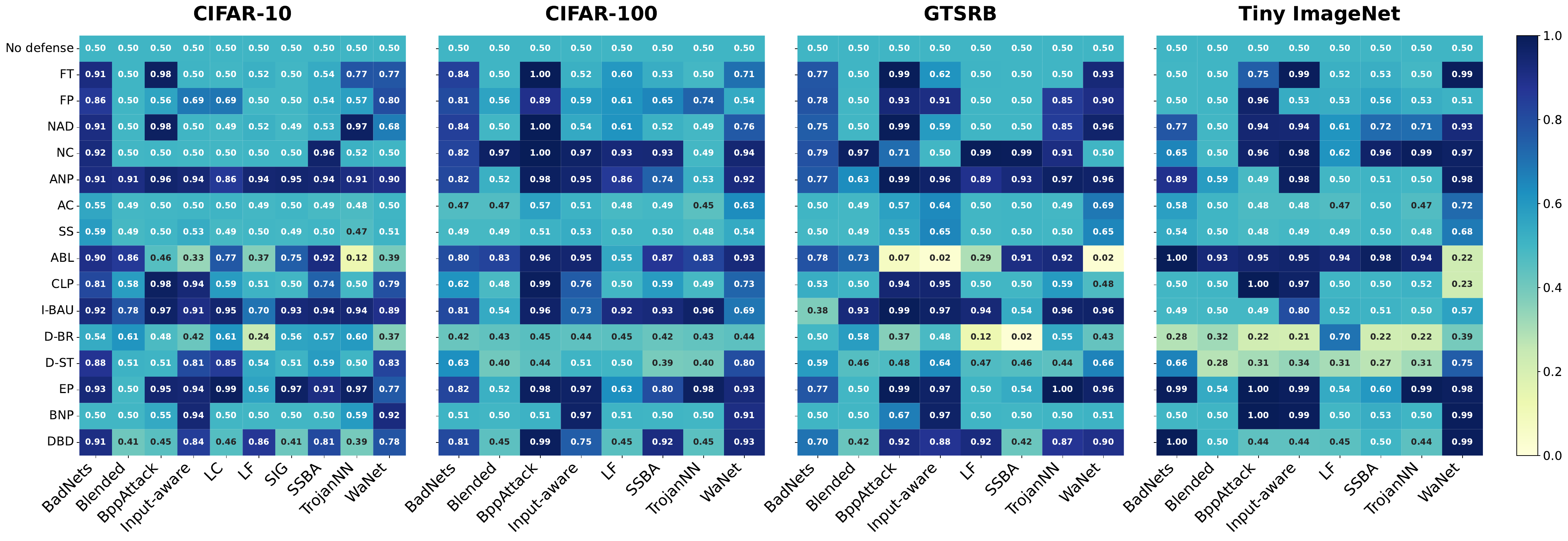}
    \vspace{-1em}
    \caption{Heat map of DER scores of different attack-defense pairs.}
    \label{fig:der}
    \vspace{-.6em}
\end{figure*}

\subsection{Results overview}

Here we overview the performance distribution of attack-against-defense pairs from three perspectives, \ie, C-Acc \vs ASR, R-Acc \vs ASR, as well as DER. For simplicity, we take the evaluations on PreAct-ResNet18 and poisoning ratio 5\% as the example, and the full results will be shown in \textbf{Supplementary Material} and the leaderboard.

\textbf{C-Acc \vs ASR of attack-defense pairs}. 
%We present the performance distribution of attack-defense pairs within a single model architecture, specifically the PreAct-ResNet18, and with a designated poisoning ratio of 5\%, as illustrated in Figure \ref{fig:ca-asr}. The performance is shown through clean accuracy (C-Acc) and attack success rate (ASR). 
The attack-against-defense performance measured by C-Acc \vs ASR is shown in Figure \ref{fig:ca-asr}. 
From the attacker's standpoint, an optimal performance means achieving high C-Acc and high ASR at the same time, corresponding to the top-right corner. Conversely, from the defender's standpoint, desirable performance involves maintaining high C-Acc and low ASR simultaneously, positioning it at the top-left corner. Most color marks are noted to be concentrated at almost the same horizontal levels with the mark of \textit{no defense}, indicating that the majority of defense methods can effectively alleviate the backdoor effect without causing a large decline in clean accuracy. 

\textbf{R-Acc \vs ASR of attack-defense pairs}. 
The attack-against-defense performance measured by R-Acc \vs ASR is shown in Figure \ref{fig:ra-asr}. 
As demonstrated in Section \ref{sec: benchmark}, it should satisfy $\text{ASR} + \text{R-Acc} \leq 1$. From the defender's standpoint, 
it is desired that the reduction in ASR is equivalent with the increase in R-Acc, which implies that the defense can not only deactivate the backdoor for the poisoned sample, but also recover its prediction to the correct label. 
The gap between $\text{ASR} + \text{R-Acc}$ and 1 can reflect the recovery capability of the defense, \ie, smaller gap indicating better capability. 
In Figure \ref{fig:ra-asr}, this gap corresponds to the perpendicular distance from one mark to the anti-diagonal line. 
It is interesting to observe that the overall distances on different datasets are ranked in ascending order as CIFAR-10 $<$ GTSRB $<$ CIFAR-100 $<$ Tiny ImageNet, which is exactly consistent with the rank of the number of candidate classes in these datasets. 
It fits our intuition that if there are more candidate classes, it is more different to recover the prediction of poisoned sample to the correct label. 
In short, this overview reveals that the recovery capability of the defense is strongly correlated with the dataset. 

% achieving an equivalent reduction in ASR and an increase in R-Acc is the objective. In other words, the aim is to restore the prediction of the poisoned sample to the correct class after the defense. Observing the colour patterns, it becomes intriguing to note their close to the anti-diagonal line. (\ie, $\text{ASR} + \text{R-Acc} = 1$) on CIFAR-10 (the first column) and GTSRB (the third column), while most patterns are from that line on CIFAR-100 (the second column) and Tiny ImageNet (the last column). We believe it is highly related to the number of classes of the dataset. Given a large number of classes, the task of recovering an accurate prediction after defense becomes very challenging. 

\textbf{DER of attack-defense pairs}. 
As shown in Figure \ref{fig:der}, we use the heat map of the DER scores to present an overall comparison between different attack and defense methods. 
From the perspective of the attacker, we can see that in general, most defense methods show high DER on BadNets, which tells that this basic attack can be easily eliminated by most defenses. 
In comparison, it is interesting to see that the Blended remains a low DER across 4 different datasets under most defenses, showing that this classic method  is still a strong baseline to evaluate the defence performance. 
%But notice that Blended use a static pattern as the trigger, which may explain part of why it works well and remind us that this attack sacrifices stealthiness for attack effectiveness. 
%
In terms of defense, we find that there is a large decline in DER for most defenses on between the three small size datasets (\ie, CIFAR-10, CIFAR-100, GTSRB) and the last large size dataset (\ie, Tiny-Imagenet). 
It implies that it is necessary to evaluate the defense performance on large size datasets in future works.

% reflects ANP as an effective defense method may still need to enhance its performance as for large size image dataset. We can also find out that for AC and Spectral, since they are relatively early stage detection methods, they are shown to be quite ineffective as for all datasets, and are not suitable for further usage. 

% These figures could provide a big picture of the performance of most attacks against defenses. Due to the space limit, the results of other settings are presented in the \textbf{Appendix}.

%% file: sections/analysis.tex
\section{Analysis}
\label{sec: analysis}

Based on the attack-against-defense evaluation results presented above, in the following we present in-depth analysis from 8 diverse perspectives, 
to reveal characteristics and mechanisms of backdoor learning. 
As summarized in Table \ref{table: analysis}, our analysis covers four major components of backdoor learning, \ie, data, model architecture, algorithm, and learning procedure, of which the analysis will be sequentially introduced below.

\input{tables/experiment}

\subsection{Effect of data}

% Here, we delve into the effect of data on both backdoor attack and defence. We pose questions, explore phenomenon and conduct analyses from four distinct perspectives regarding the data: (1) \textit{the influence of different poisoned samples}, (2) \textit{the effect of varying poisoning ratios}, (3) \textit{the generalization capability of backdoor triggers}, and (4) \textit{the stealthiness of backdoor triggers}.

Here, we delve into the effect of data on both backdoor attack and defence, including 4 data related perspectives: (1) \textit{the influence of different poisoned samples}, (2) \textit{the effect of varying poisoning ratios}, (3) \textit{the generalization capability of backdoor triggers}, and (4) \textit{the stealthiness of backdoor triggers}.

\input{sections/analysis/sample_selection}
\input{sections/analysis/poisoning_ratio}

\input{sections/analysis/trigger_generalization}

\input{sections/analysis/stealthiness}
% % shaokui

\subsection{Effect of model}

In this section, we delve into backdoor learning from the model perspective, covering \textit{the influence of model architecture}, and \textit{the sharpness of backdoored model}.

% Here, we delve into the impact of the model structure on both backdoor attacks and defence. We pose questions, explore phenomena and conduct analyses from two distinct perspectives regarding the model structure: (1) \textbf{the influence of different model structure}, and (2) \textbf{the sharpness of backdoored model}.

% Here, we delve into the impact of data on both backdoor attacks and defence. We pose questions, explore phenomena and conduct analyses from four distinct perspectives regarding the data: (1) \textbf{the influence of different backdoor samples}, (2) \textbf{the effects of varying poisoning ratios}, (3) \textbf{the generalization capability of poisoning triggers}, and (4) \textbf{the stealthiness of poisoning triggers}.

% mingda
\input{sections/analysis/architecture}

\input{sections/analysis/sharpness}

% mingli 

\subsection{Effect of attack/defense algorithm}

Here we analyze the sensitivity of each individual backdoor attack/defense algorithm to their critical hyper-parameters, such that we have a better understanding about their mechanisms and practical performance. 
Our analysis for all algorithms in this section are based on the evaluations on the CIFAR-10 dataset with Preact-ResNet18. 

% We discuss the \textbf{sensitivity} of different backdoor attack and defense algorithms to analyze their mechanisms.

\input{sections/analysis/sensitivity}

\subsection{Effect of backdoor learning}

We investigate the backdoor learning mechanism by observing \textbf{the difference on the training process} between clean samples and poisoned samples, from the following two perspectives. 

\input{sections/analysis/quick_learning_forget}

%% file: tables/experiment.tex
\begin{table*}[]
\renewcommand\arraystretch{1.5}
\center
\vspace{-1em}
\caption{Summary of analysis of backdoor learning. }
%\vspace{-1em}
\label{table: analysis}
\scalebox{0.9}{
\begin{tabular}{m{.09\textwidth}<{\centering} 
 m{.47\textwidth} m{.2\textwidth}<{\centering}  m{.25\textwidth} } %{clcc}
\hline
\multirow{2}{*}{\textbf{Components}}  & \multicolumn{1}{c}{\multirow{2}{*}{\textbf{Contents}}}                                             & \multicolumn{2}{c}{\textbf{Tools}}               \\ \cline{3-4} 
                           & \multicolumn{1}{c}{}                                                                     & \textbf{Evaluation} & \textbf{Analysis tools}                   \\ \hline
\multirow{4}{*}{Data}      & 1) How do different poisoned samples affect the backdoor effect?                          & ASR        &   Poisoned sample selection strategy                         \\ \cline{2-4} 
                           & 2) What is the effect of poisoning ratios on backdoor attack and defense?      & ASR        & Activation/t-SNE           \\ \cline{2-4} 
                           & 3) Whether the trigger has the ability to generalize?                                     & ASR        & Activation path  \& Trigger strength         \\ \cline{2-4} 
                           & 4) What is the stealthiness of the trigger for different backdoor attacks? &  Seven image quality metrics          & Image quality assessment   \\ \hline
\multirow{2}{*}{Model} & 1) What is the effect of model architecture on different algorithms?                       & ASR        & Activation path            \\ \cline{2-4} 
                           & 2) What is the sharpness of the loss landscape for backdoored model?                         &  Eigenvalue density  &  Hessian matrix \&  Activation ratio   \\ \hline
\multirow{2}{*}{Algorithm} & 1) What is the sensitivity to hyper-parameters for each attack algorithm?                                           & ASR  &  Trigger strength                          \\
\cline{2-4} 
                           &  2) What is the sensitivity to hyper-parameters for each defense algorithm?                                           & ASR  &  Trigger strength 
                            \\
\hline
\multirow{2}{*}{Learning}         & 1) What is the difference on learning speed between poisoned and clean samples?    & Training loss curve      & GSNR \& Gradient norms \\ 
\cline{2-4} 
                          & 2) What is the difference on memorization between poisoned and clean samples? & Forgetting ratio & GSNR \& Gradient norms
                          \\
\hline
\end{tabular}
}
\vspace{-.5em}
\end{table*}

%% file: sections/analysis/sample_selection.tex
\paragraph{The influence of different poisoned samples}
\label{sec: subsec effect of backdoor sample selection}

\comment{hongrui chen
角度：不同dataset、不同model、不同ratio对4个selection顺序的影响。
还差一些图像。}

\begin{figure*}[htbp]
    % \subfigure[Selection]{
        \begin{minipage}[t]{0.95\linewidth}
        \centering
            \includegraphics[width=\textwidth]{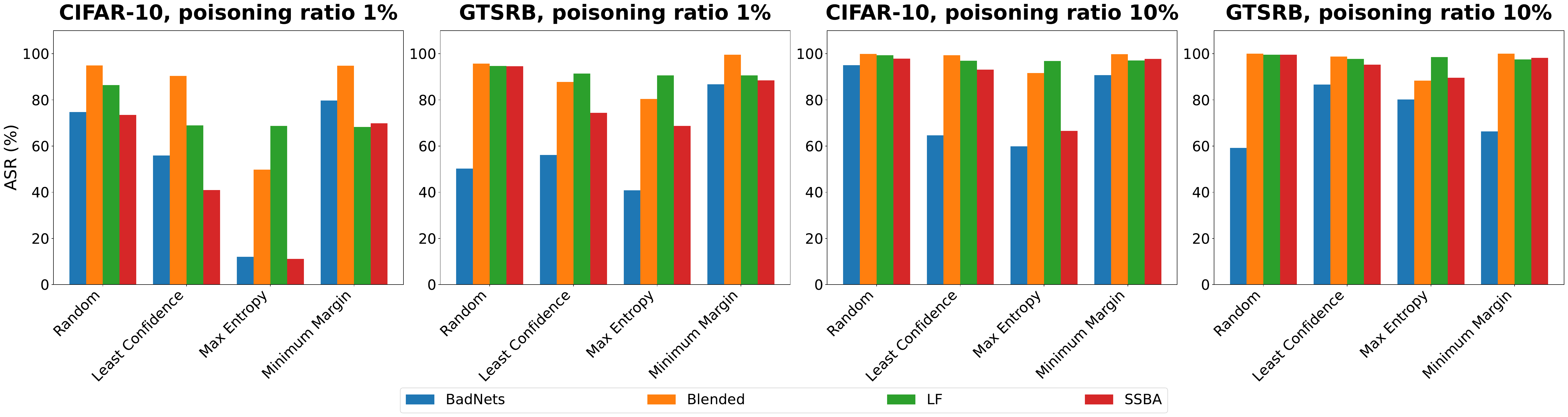}
        \end{minipage}%
    % }%
    \centering
    \vspace{-1em}
    % \caption{ASRs for different attack methods on different datasets with different poisoning ratios concerning different sample selection strategies.}
    \caption{The influence of different selection strategies of poisoned samples on the attack performance measured by ASR. 
    % The horizontal labels represent four strategies of poisoned sample selection.
    %The labels of the horizontal axis represent four strategies of poisoned sample selection.
    }
    \label{fig: selection}
    \vspace{-1.2em}
\end{figure*}

In most existing backdoor learning works, a widely adopted setting is that some clean samples with the given poisoning ratio are randomly selected from the whole clean dataset to generate poisoned samples. Consequently, it is often observed that there is large fluctuation of the attack performance when we repeat the attack evaluation. 
To study the influence of different poisoned samples, here we compare the random strategy with another three strategies. 
Specifically, we firstly train a model on the whole clean training dataset, and then record the prediction confidences \wrt all classes of each training sample by this model. 
The \textit{least confidence strategy} means that the samples of the top-$r\%$ ($r$ indicating poisoning ratio) least confidence \wrt their ground-truth labels are selected for poisoning. 
The \textit{maximum entropy strategy} means that the samples of the top-$r\%$ maximum entropy of the prediction confidences \wrt all classes are selected. 
The \textit{minimum margin strategy} means that the samples of the top-$r\%$ minimum margin between the largest and the second largest confidence are selected.

Our analysis is based on the evaluations of two datasets (\ie, CIFAR-10, GTSRB), one model (\ie, PreAct-ResNet18), two poisoning ratios (10\% and 1\%), and four attack methods (BadNets, Blended, LF, and SSBA). 
%The results is shown in Figure \ref{fig: selection}, with four rows corresponding to four attack methods and four columns corresponding to two datasets with two poisoning ratios. 
As shown in Figure \ref{fig: selection}, the random and minimum margin strategies generally outperform the remaining two strategies at the poisoning ratio 1$\%$, while most strategies show similar performance at the poisoning ratio 10$\%$. 
It demonstrates that the selection of poisoned samples is an important factor for attack, especially at the low poisoning ratio. However, how to find a more effective and stable selection strategy than the random strategy is still an open problem, which deserves to be further explored.

%% file: sections/analysis/poisoning_ratio.tex
\begin{figure*}[htpb]
    \centering
    \includegraphics[width=0.9\textwidth]{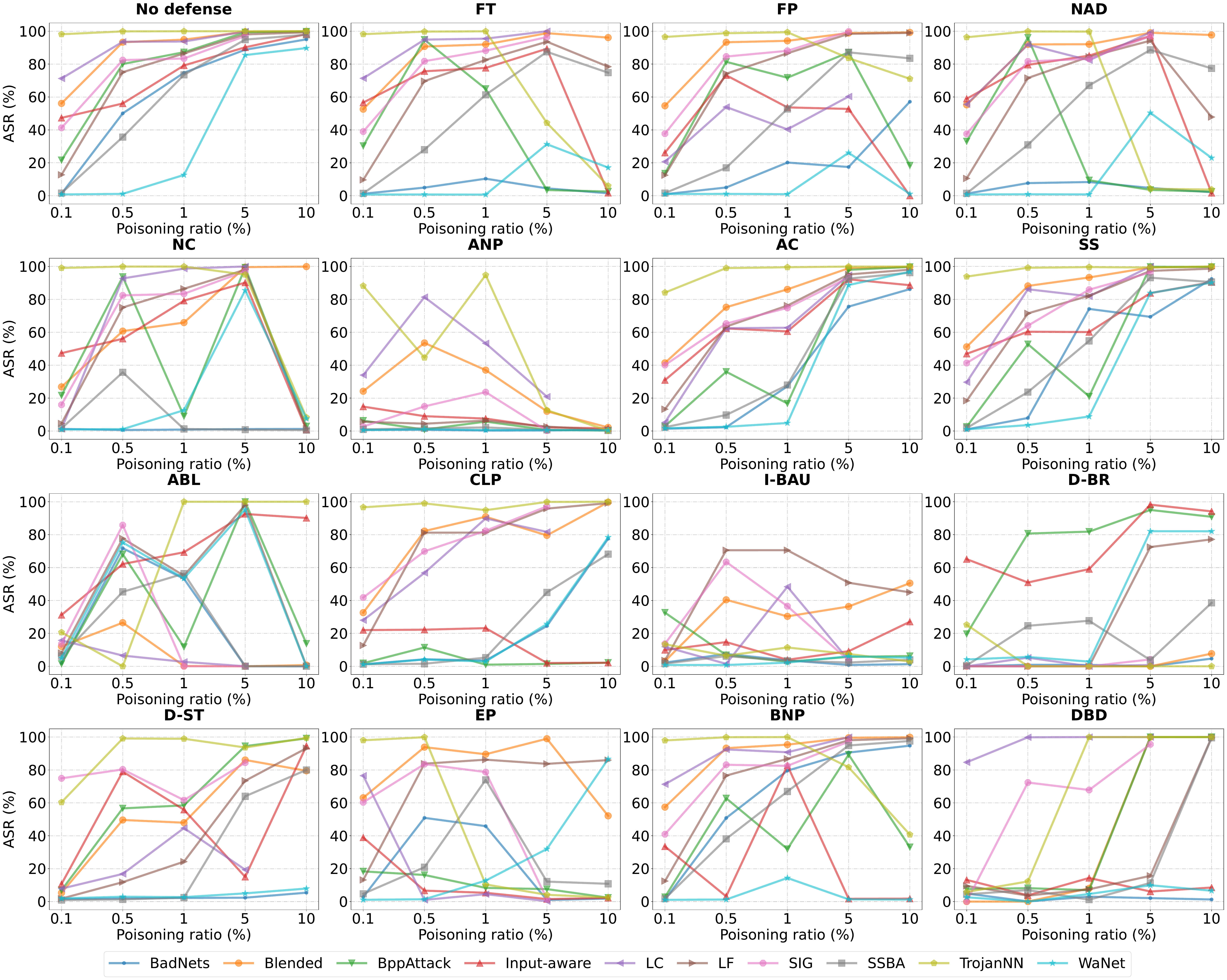}
    \vspace{-1em}
    \caption{The effects of different poisoning ratios on backdoor attack and defense performance measured by ASR.}
    \vspace{-1em}
    \label{2-attack-asr-preact}
\end{figure*}

\paragraph{Effect of varying poisoning ratio}
\label{sec: subsec effect of poisoning ratio}

Although it is supposed that a higher poisoning ratio of attack methods should lead to higher ASR, we would like to investigate that whether higher poisoning ratio can still lead to higher ASR under different defenses. Specifically, we study the effect of the poisoning ratio on backdoor attack performance under different attack-defense pairs. 
Figure ~\ref{2-attack-asr-preact} visualizes the results on CIFAR-10 and PreAct-ResNet18 \wrt each poisoning ratio (including 0.1\%, 0.5\%, 1\%, 5\%, 10\%) for all attack-defense pairs, and each sub-figure corresponds to one defense method. 
According to the trend of each defense method, we categorize our analysis into the following groups: 
% For a better understanding of the trend of defense methods, we carry out the following deeper analysis:

% In sub-figures (1,6,7)\red{Need Update}, ASR curves increase in most cases, being consistent with our initial impression that higher poisoning ratios lead to stronger attack performance. 
% However, in other sub-figures, there are surprisingly sharp drops in ASR curves. To understand such \textit{abnormal} phenomenon, we conduct deep analysis for these defenses, as follows.  

\begin{figure*}
     \centering
     \includegraphics[width=.9\textwidth]{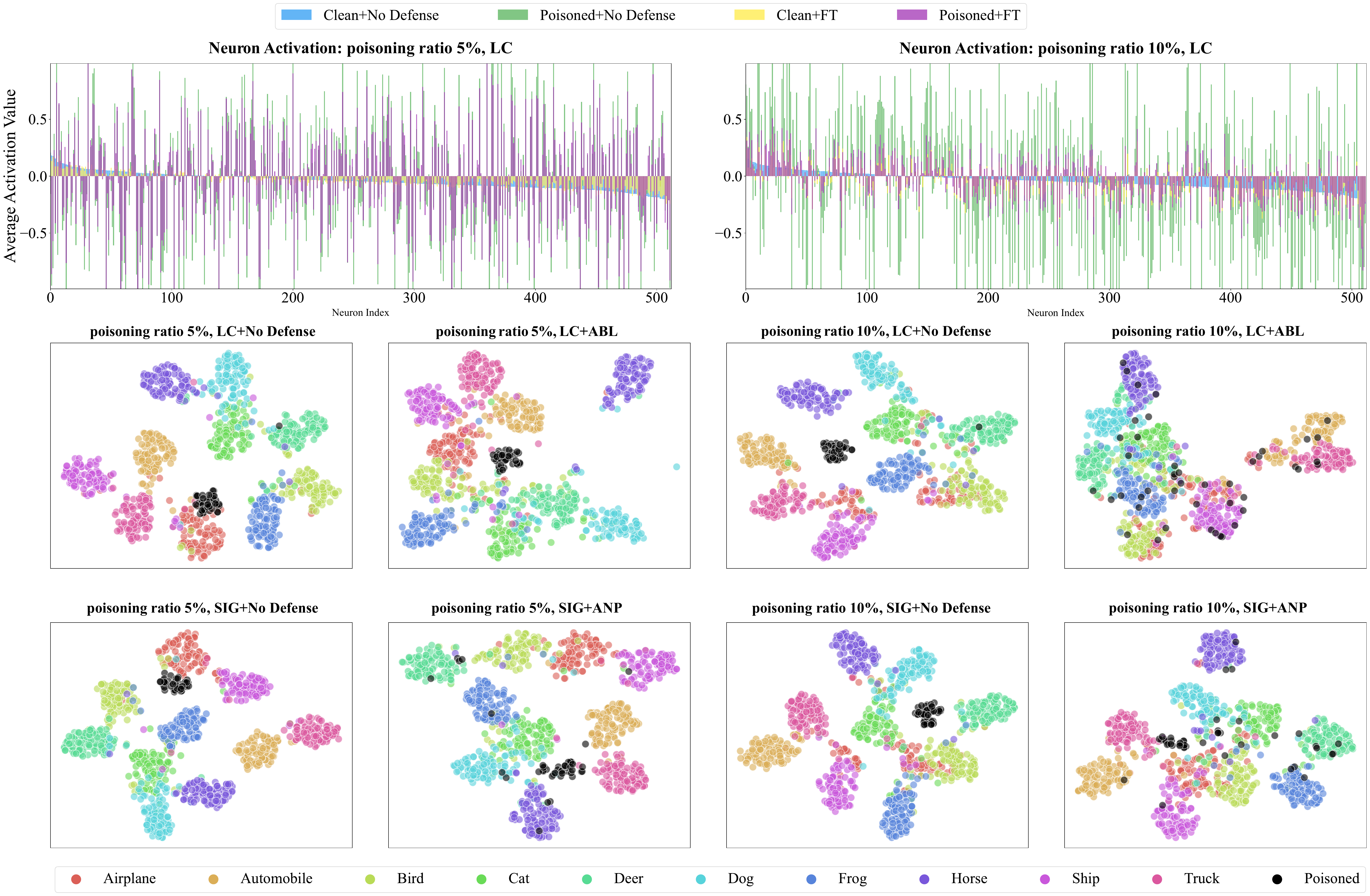}
     \vspace{-1em}
     \caption{The changes of neuron activation values due to the FT defense (\textbf{Top row}), and the changes of t-SNE visualization of feature representations due to the ABL defense (\textbf{Middle row}) and the ANP defense (\textbf{Bottom row}), respectively. }
    \label{fig:abnormal}
    \vspace{-1.5em}
\end{figure*}

\textbf{Analysis of No defense/AC/SS/CLP/D-BR/D-ST/DBD}.  
In sub-figures of No defense, AC \cite{ac}, SS \cite{tran2018spectral}, CLP \cite{clp-2022}, D-BR \cite{chen2022dbr}, D-ST \cite{chen2022dbr}, and DBD\cite{huang2022backdoor}, the ASR curves of these methods, exhibit a monotonically increasing trend in most cases. 
% , aligning with our initial expectation. 
Except CLP, all remaining defense methods involves poisoned sample filtering. Consequently, given the higher poisoning ratio, it is likely to miss more poisoned samples, leading to higher ASR. 
CLP is a data-free pruning-based method and mainly utilizes a newly defined metric of neuron weights to prune backdoor related neurons. Given the higher poisoning ratio, it is likely that more neurons are related to backdoor effect, causing that it is more difficult to prune most backdoor related neurons while keeping the clean accuracy. 

% These methods mainly include sample segmentation, the more backdoor samples, the more backdoor samples missed, resulting in higher ASR after defense. 
% Since CLP does not use any clean samples, it is itself affected by the poisoning effect of the backdoor model, and the higher the poisoning ratio, the more difficult the defense. However, the rest of the defense methods exhibit greater complexity, even surprisingly sharp drops in ASR curves. A more detailed explanation shows that other defenses do not follow a simple increasing rule. 

% We can get some clues about the trend from the property of defensive methods. 

% The two iterative pruning/fine-tuning methods, ANP and I-BAU, perform great along all five poisoning ratios, except on a few attack methods. For ANP, only TrojanNN and Blended remain at a high ASR under a low poisoning ratio. For I-BAU, only blended with two frequency-based methods can maintain over 40\% ASR under part of the poisoning ratio. This fact shows that iterative methods can work well under most poisoning ratios due to their progressive mitigation of inserted backdoors.

\textbf{Analysis of FT/FP/NAD/NC}. 
FT, FP \cite{fp}, NAD \cite{nad-iclr-2020} and NC\cite{wang2019neural} are all fune-tuning based defense methods, \ie, fine-tuning on a small subset of clean data (\ie, $5\%$ training data). The trend of their ASR curves under different defense methods look similar, \ie, firstly increasing then decreasing sharply along with the increasing of the poisoning ratio. 
We take FT as the example for a deep analysis. 
As shown in Figure \ref{fig:abnormal}, we compare the performance of the LC attack between the poisoning ratio $5\%$ and $10\%$. 
We first investigate the changes in the average neuron activation before and after the defense. As shown in the top row, the changes between \textit{Poisoned+No Defense} (green) and \textit{Poisoned+FT} (purple) in the case of $5\%$ are much smaller than those in the case of $10\%$. It implies that the backdoor effect could be significantly mitigated by FT when the poisoning ratio is high enough. 
We think on possible reason is that when the poisoning ratio is not very high (\eg, $5\%$), the model fits clean samples very well, thus fine-tuning on a subset of clean sample will not significantly change the model weight. In contrast, in the case of high poisoning ratio, the fitting to clean samples is no longer very well, the clean fine-tuning may change the model weight a lot. 
To verify that, we record that the clean accuracy on the $5\%$ clean data used for fine-tuning by the backdoored model before defense is $99\%$ in the case of $5\%$ poisoning ratio, while $92\%$ in the case of $10\%$ poisoning ratio. It explains why their changes in neuron activation values are different. 
%
%We believe that above analysis could explain why fine-tuning based methods have larger ASR values remaining at medium-level poison ratios, whose curves increase first and then decrease at higher poison ratios.

\textbf{Analysis of ANP/I-BAU}.  
The two iterative pruning/fine-tuning methods, ANP\cite{wu2021adversarial} and I-BAU\cite{i-bau}, perform relatively well at all five poisoning ratios, except on a few attacks. ANP prunes the neurons that are sensitive to the adversarial neuron perturbation, and I-BAU unlearns the adversarial samples, which is probably poisoned samples. 
Given a high poisoning ratio, the backdoor effect is supposed to be more pronounced, thus it is easier to identify backdoor related neurons and samples. 
% We find that when the poisoning ratio is high, more neurons will be pruned by ANP. Thus, the ASR may decrease. 
For example, given the SIG \cite{SIG} attack, the pruned neurons by ANP are 328 and 466 for $5\%$ and $10\%$ poisoning ratios, respectively. As shown in the last row of Figure \ref{fig:abnormal}, poisoned samples still gather together for $5\%$, while separated for $10\%$. This phenomenon explains why some attack methods have a higher ASR for medium poison ratios than others.

% NC, NAD, FT, and FP are mainly fine-tune-based methods. Since they all fine-tuned on a small subset of clean data (\ie, $5\%$ training data) and do not iterative fine-tune/prune, their performance is highly relied on given a clean dataset quality; they may have unstable performance and no clear trend along different poisoning ratio.

% Since EP, BNP, and MBNS rely on the statistical information from mini-batch, they may be influenced by noise in mini-batch and cannot have a stable trend along different poisoning ratios.

\textbf{Analysis of ABL}. 
The ABL \cite{li2021anti} method uses the loss gap between the poisoned and clean samples in the early training period to isolate some poisoned samples. We find that the loss gap in the case of a high poisoning ratio is larger than that in the case of a low poisoning ratio. Take the LC \cite{turner2019labelconsistent} attack on CIFAR-10 as an example. In the case of 5$\%$ poisoning ratio, the isolated 500 samples by ABL are 0 poisoned and 500 clean samples, such that the backdoor effect cannot be mitigated in later backdoor unlearning in ABL. In contrast, the isolated 500 samples are all poisoned in the case of 10$\%$ poisoning ratio. The t-SNE visualizations shown in the second row of Figure \ref{fig:abnormal} also verify this point.  
% This fact explains why at 10\%, ABL can perform better compared to 5\%. \red{TODO} We may notice that there is another peak at 0.5\%. 

\textbf{Analysis of EP and BNP}. 
Both EP \cite{ep-bnp-2022} and BNP \cite{ep-bnp-2022} aim to prune backdoor related neurons, which are identified according to the different distributions of the activation on clean and poisoned samples between benign and backdoor related neurons. They show diverse defense performance against different attacks. It implies that the utilized activation distribution may vary significantly under different poisoning ratios and different triggers, leading to the variation of the pruning-based defense performance.

\textbf{In summary}, the above analysis demonstrates that an attack with higher poisoning ratios doesn't mean better attack performance, and it may be easier to be defended. The general reason is that attack with higher poisoning ratios will amplify the difference between poisoned and clean samples, which could be detected and utilized by defender. It inspires two challenges that deserve further exploration: \textit{how to achieve the desired attack performance using fewer poisoned samples, and how to defend weak attacks with low poisoning ratios}. 
% Moreover, considering the randomness due to weight initialization and some methods' mechanisms, we repeat the above evaluations several times. Although some fluctuations occur, the trend of ASR curves is similar to that in Figure \ref{2-attack-asr-preact}. More details and analysis are presented in  \textbf{Supplementary Material}. \comment{We have also done correlation analysis for ASR after AC with TPR of detection results in the first stage of AC and ASR after NC with L1-norm of the reverse-engineered trigger of NC, but no strong correlation is found. }

% \textbf{Analysis of ANP/I-BAU} 
% The ANP \cite{wu2021adversarial} prunes the neurons that are sensitive to the adversarial neuron perturbation by setting a threshold. As suggested in \cite{wu2021adversarial}, this threshold is fixed as 0.2 in our evaluations. We find that when the poisoning ratio is high, more neurons will be pruned. Thus, the ASR may decrease. For example, given the SIG \cite{SIG} attack, the pruned neurons by ANP are 328 and 466 for $5\%$ and $10\%$ poisoning ratios, respectively. 
% As shown in the last row of Figure \ref{fig:abnormal}, poisoned samples still gather together for $5\%$, while separated for $10\%$. This phenomenon explains why some attack methods have a higher ASR for medium poison ratios than others.

%% file: sections/analysis/trigger_generalization.tex
\paragraph{Trigger generalization}
\label{sec: trigger generalization}

\begin{figure*}[htbp]
\centering
\subfigure[BadNets, PreAct-ResNet18]{
\begin{minipage}[t]{0.5\linewidth}
\centering
\includegraphics[width=\textwidth]{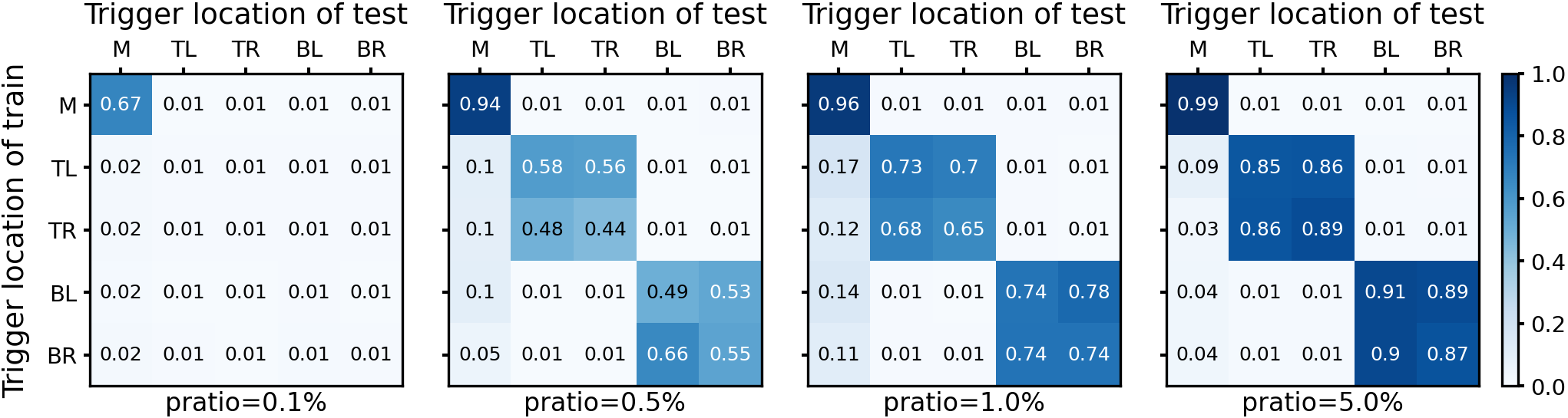}
\vspace{-.6em}
\end{minipage}%
  }%
\subfigure[BadNets, VGG19-BN]{
\begin{minipage}[t]{0.5\linewidth}
\centering
\includegraphics[width=\textwidth]{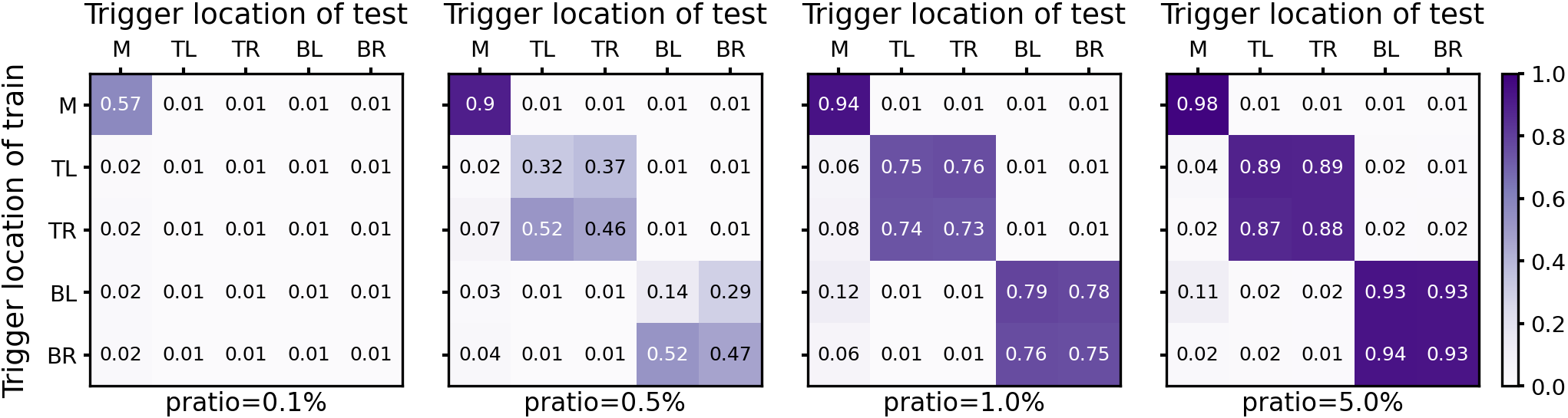}
\vspace{-.6em}
\end{minipage}%
}%

\subfigure[Blended, PreAct-ResNet18]{
\begin{minipage}[t]{0.5\linewidth}
\centering
\includegraphics[width=\textwidth]{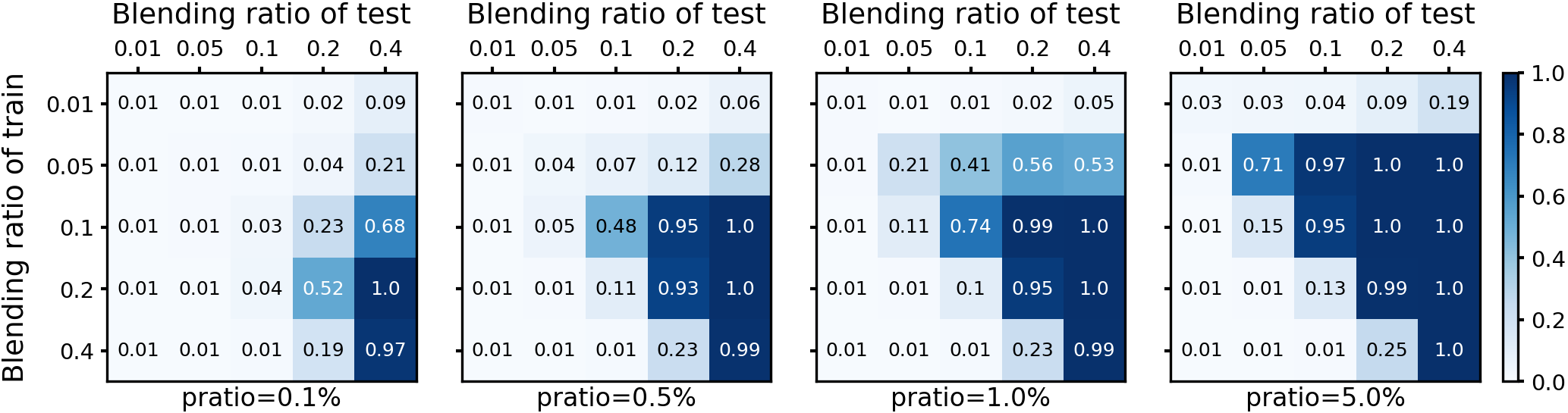}
\vspace{-.6em}
\end{minipage}
}%
\subfigure[Blended, VGG19-BN]{
\begin{minipage}[t]{0.5\linewidth}
\centering
\includegraphics[width=\textwidth]{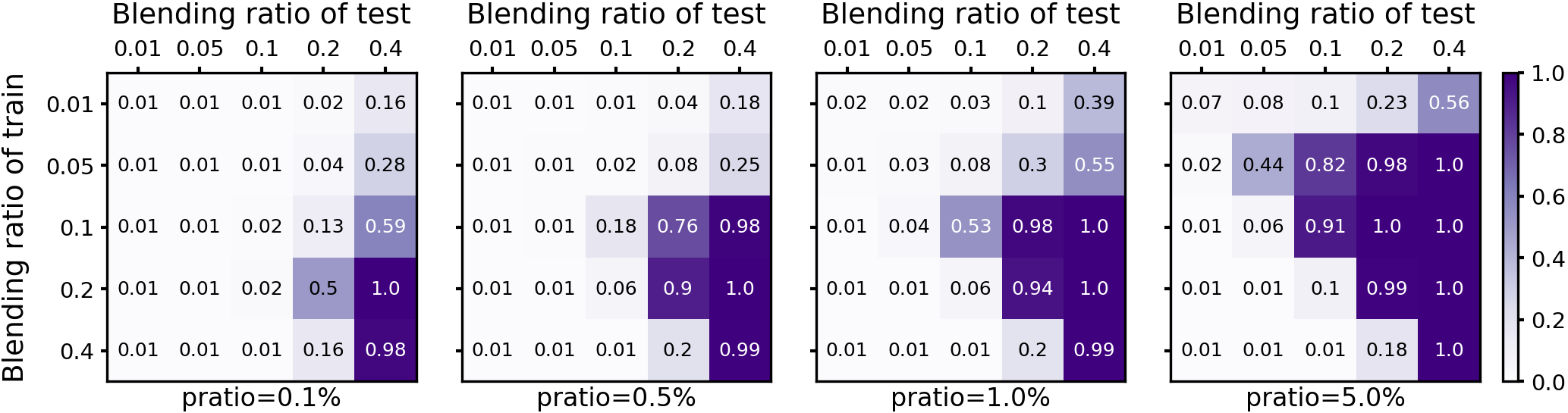}
\vspace{-.6em}
\end{minipage}
}%

\subfigure[SIG, PreAct-ResNet18]{
\vspace{-.6em}
\begin{minipage}[t]{0.5\linewidth}
\centering
\includegraphics[width=\textwidth]{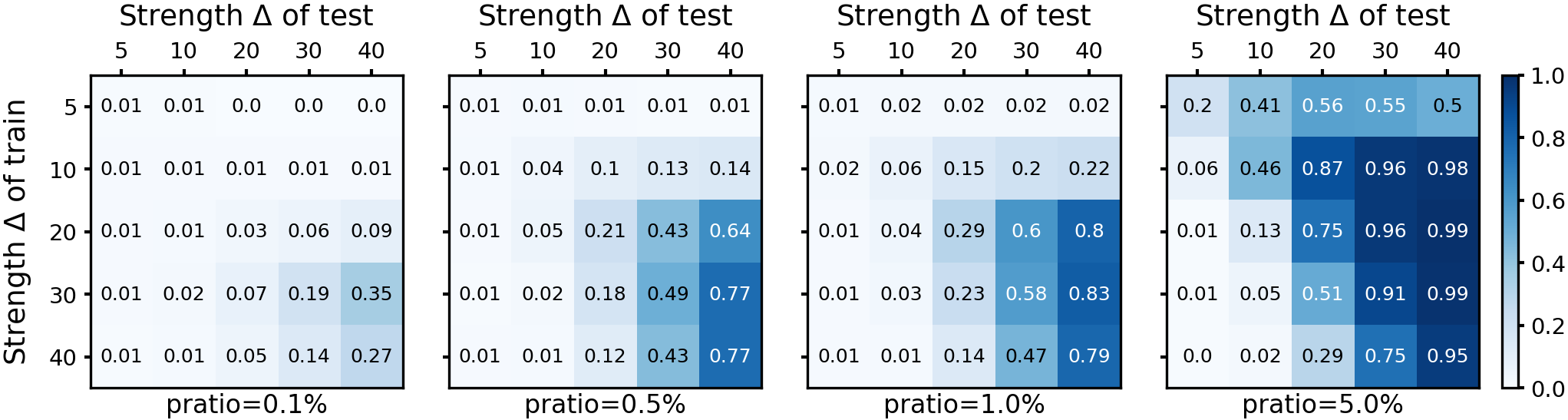}
\vspace{-.6em}
\end{minipage}
}%
\subfigure[SIG, VGG19-BN]{
\vspace{-.6em}
\begin{minipage}[t]{0.5\linewidth}
\centering
\includegraphics[width=\textwidth]{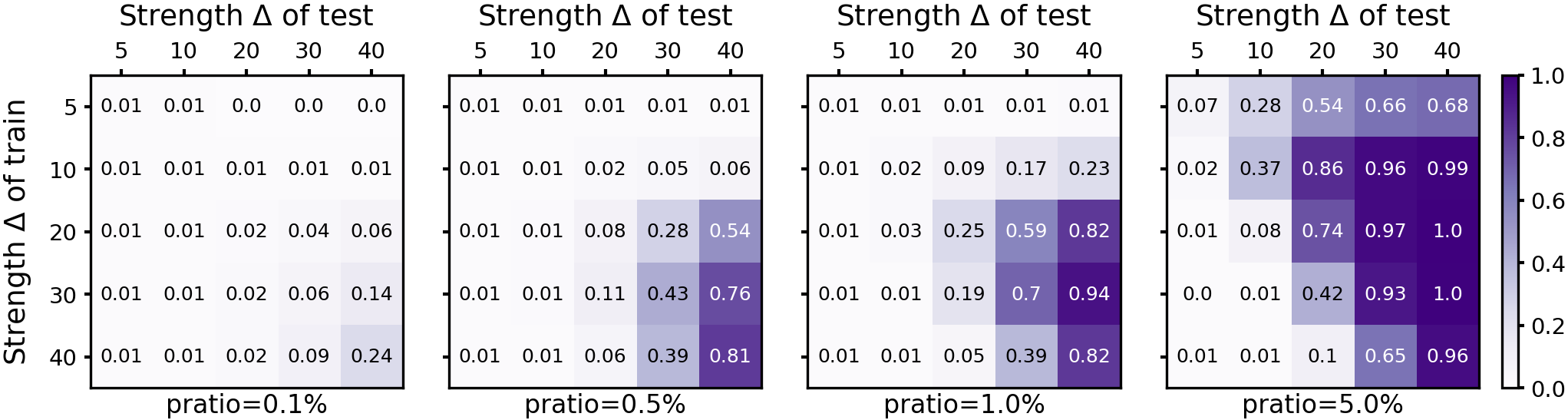}
\vspace{-.6em}
\end{minipage}
}%

\subfigure[Analysis of BadNets, PreAct-ResNet18]{
\begin{minipage}[t]{0.5\linewidth}
\centering
\includegraphics[width=\textwidth]{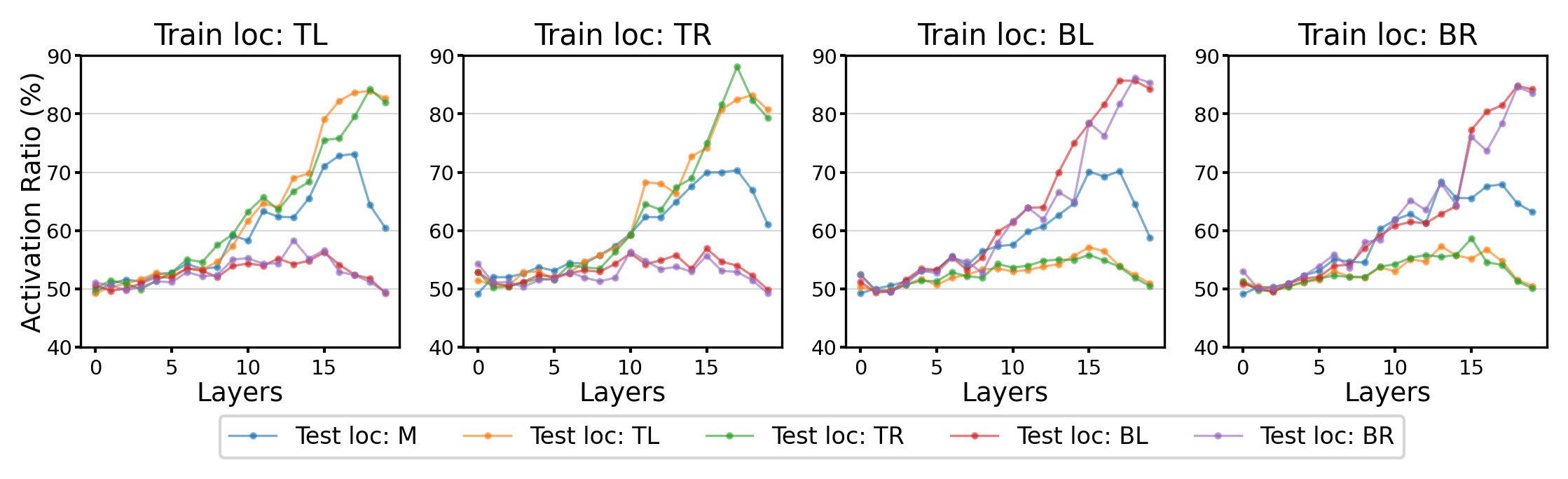}
\vspace{-.6em}
\end{minipage}
}%
\subfigure[Analysis of Blended, VGG19-BN]{
\begin{minipage}[t]{0.5\linewidth}
\centering
\includegraphics[width=\textwidth]{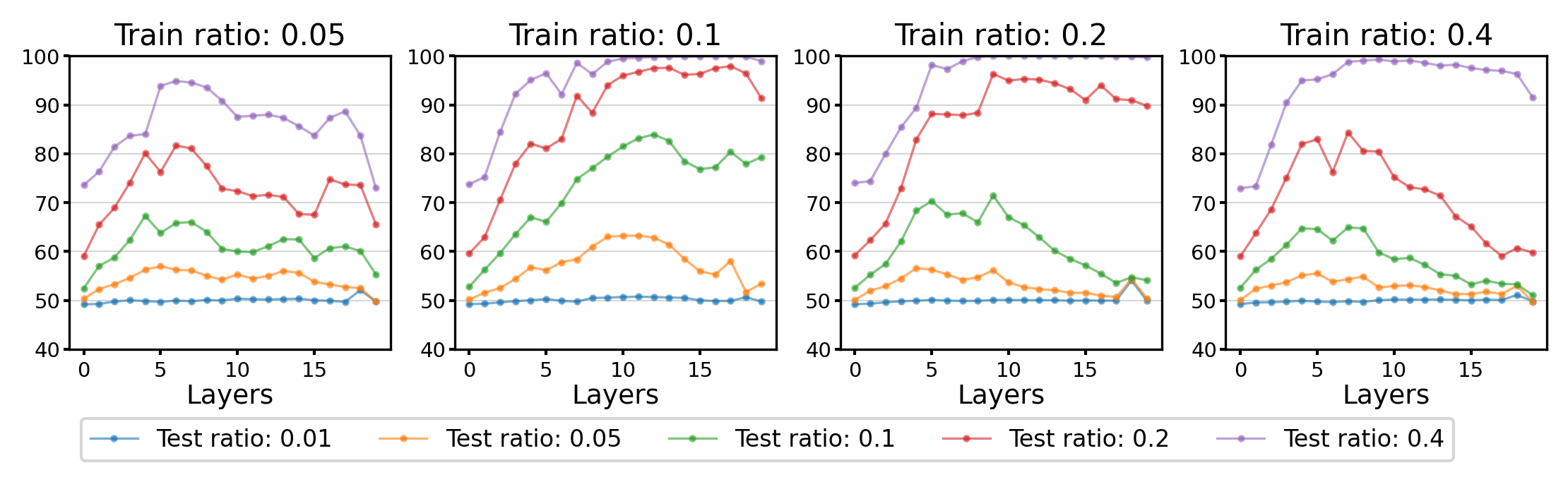}
\vspace{-.6em}
\end{minipage}
}%
\centering
\vspace{-.6em}
\caption{Trigger generalization results on CIFAR-10  with \textcolor{darkblue}{\textbf{PreAct-ResNet18}} and \textcolor{darkpurple}{\textbf{VGG19-BN}}. The vertical axis represents the parameters of the training phase, and the horizontal axis represents the parameters of the testing phase. ASR is marked in each box and the color of the box represents the level of ASR.}
\label{fig: trigger_generalization_cifar10}
\vspace{-1em}
\end{figure*}

In most data-poisoning based backdoor attacks, it is commonly assumed that the triggers used for both training and testing are same. Here we aim to explore what will happen if there is difference between training and testing triggers, \ie, the trigger generalization in backdoor attack. 
% In this section, we aim to explore the generalization of triggers, $i.e.$, whether the malicious effectiveness of poisoned backdoor models is maintained when the triggers differ between the training and testing stages. 
%
We choose three representative data poisoning-based backdoor attacks for the evaluation of trigger generalization. Specifically, for BadNets~\cite{gu2019badnets}, we modify the trigger's position, by placing it at the middle (M), top left corner (TL), top right corner (TR), bottom left corner (BL) and bottom right corner (BR), respectively. 
For Blended~\cite{chen2017targeted}, we adjust the blending ratio, by setting it to 0.01, 0.05, 0.1, 0.2 and 0.4, respectively. 
For SIG~\cite{SIG}, we adjust the strength $\Delta$ of signal, with values of 5, 10, 20, 30 and 40, respectively. 

The evaluations of PreAct-ResNet18 and VGG19 on CIFAR-10 are shown in Figure~\ref{fig: trigger_generalization_cifar10}. The results reveal significant variations on the performance of different triggers across datasets, models, and poisoning ratios. Overall, trigger generalization improves as the poisoning ratio increases. For BadNets~\cite{gu2019badnets} on CIFAR-10, triggers in the same row are more easily activated. For example, the trigger in the top-left (TL) corner can be activated by the trigger in the top-right (TR) corner, but not by the triggers in the bottom-left (BL) or bottom-right (BR) corners. However, this phenomenon disappears on the dataset CIFAR-100. For Blended~\cite{chen2017targeted} and SIG~\cite{SIG} attacks, stronger triggers are more likely to be activated when the trigger strength exceeds a certain threshold (\eg, for Blended, pratio=0.05, blending ratio > 0.05), while weaker triggers are less likely to activate the backdoor. Additionally, stronger triggers exhibit higher ASRs during the testing phase, implying that if models are capable of learning backdoor from weaker triggers during the training stage, they will possess better trigger generalization. % 
%, and TrojanNN~\cite{liu2018trojaning} attacks, 

To better understand above observations of trigger generalization, we present the activation ratio of neurons at each layer of the neural network on poisoned samples, when using different triggers. As shown in Figure~\ref{fig: trigger_generalization_cifar10}(g), we find that the activation ratio is negatively related to the location distance between training and testing triggers, which is consistent with the above observation on trigger generation of BadNets.  
As shown in Figure~\ref{fig: trigger_generalization_cifar10}(h), we find: the activation ratio is positively related to the testing blending ratio; when the training blending ratio is high, the activation ratio gap between high and low testing blending ratios is large, and the activation gap of low testing blending ratio tend to decrease quickly along with the layer, which is consistent with the observation on trigger generalization. 
This analysis reveals that the activation ratio of neurons on poisoned samples is a good sign for trigger generalization, which deserves further exploration.

%% file: sections/analysis/stealthiness.tex
\paragraph{The stealthiness of backdoor attack}
\label{sec: The stealthiness of backdoor attack}

In addition to the attack effectiveness which can be measured by ASR, the stealthiness of poisoned samples is another desired characteristic of a good backdoor attack. Here, we analyze the stealthiness of different attacks by quantifying the visual difference between poisoned and clean samples using various image quality metrics. 
Specifically, we adopt the structural similarity index measure (SSIM) \cite{SSIM} to measure the structural similarity between each poisoned sample and its corresponding clean sample, the Frechet inception distance (FID) \cite{FID} to measure the distance between the distributions of poison samples and target-class clean samples, and the
blind/referenceless image spatial quality evaluator (BRISQUE) \cite{brisque} to measure the quality of each individual sample. 
Besides, we adopt a recent work \cite{clip_iqa} that utilizes CLIP models to compare poisoned and clean samples from diverse angles, including \textit{quality}, \textit{noisiness}, \textit{naturalness}, and \textit{reality}. 
Note that the smaller values of FID and BRISQUE indicate higher image quality, while the opposite of all others. 

\begin{figure*}[htbp]
\centering
\includegraphics[width=\textwidth]{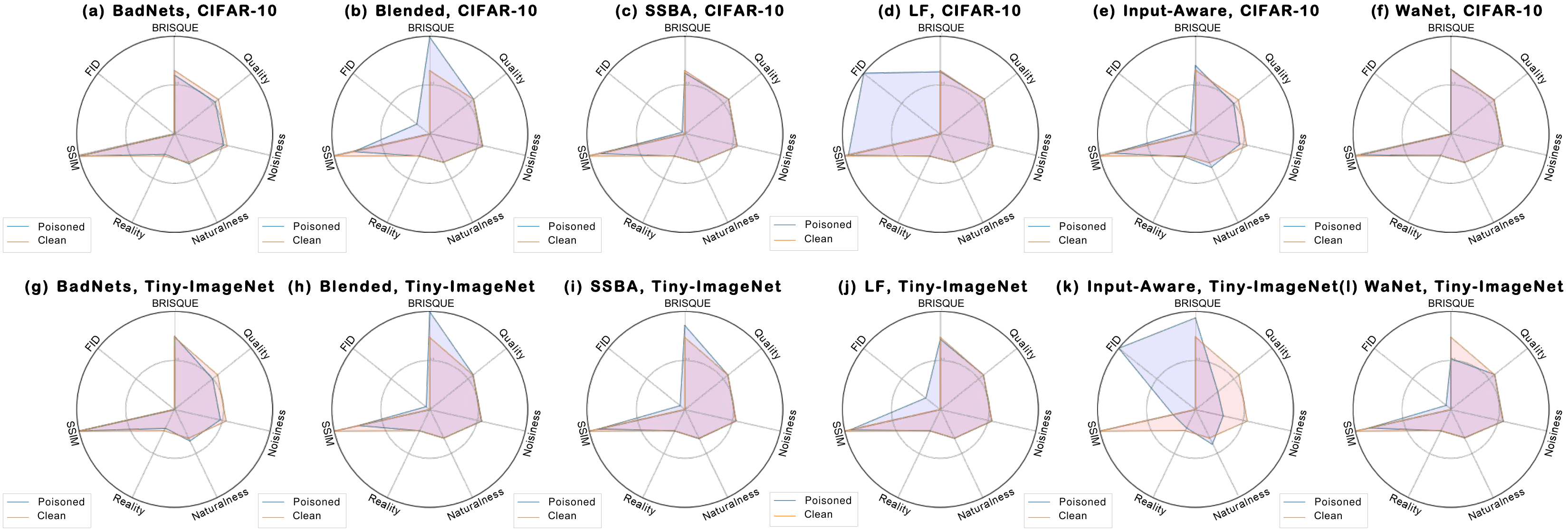}
\caption{Stealthiness analysis by comparing the image quality between poisoned and clean samples.}
\label{fig:st}
\vspace{-1em}
\end{figure*}

As shown in Figure~\ref{fig:st}, the quality difference between poisoned and clean samples is small at most cases, demonstrating the good stealthiness of most attacks. 
There are also a few special cases. 
For example, the trigger of LF attack is obtained via optimization by restricting it in the low frequency domain, while there is no constraint on the difference between the poisoned sample with trigger and the original clean sample. Thus, it may lead to large difference on FID metric. 
The trigger of Input-Aware attack is generated by a learned generative network, and there is no special constraint on the trigger or the generative network, such that the trigger is often highly visible, leading to large quality difference between poisoned and clean samples.

\comment{
Exploring CLIP for Assessing the Look and Feel of Images (AAAI 2023)
'brisque': Lower values of score reflect better perceptual quality of image, 越小越好
'quality', 越大质量越好
'noisiness',  越大越干净
'natural', 越大越自然
'real', 越大越真
 'SSIM', 越大越像
 'FID' 越小越分布越接近
}

%% file: sections/analysis/architecture.tex
\paragraph{The influence of different
model architectures}

\begin{figure}[hb]
\centering
\begin{minipage}[t]{1.0\linewidth}
\centering
\includegraphics[width=\textwidth]{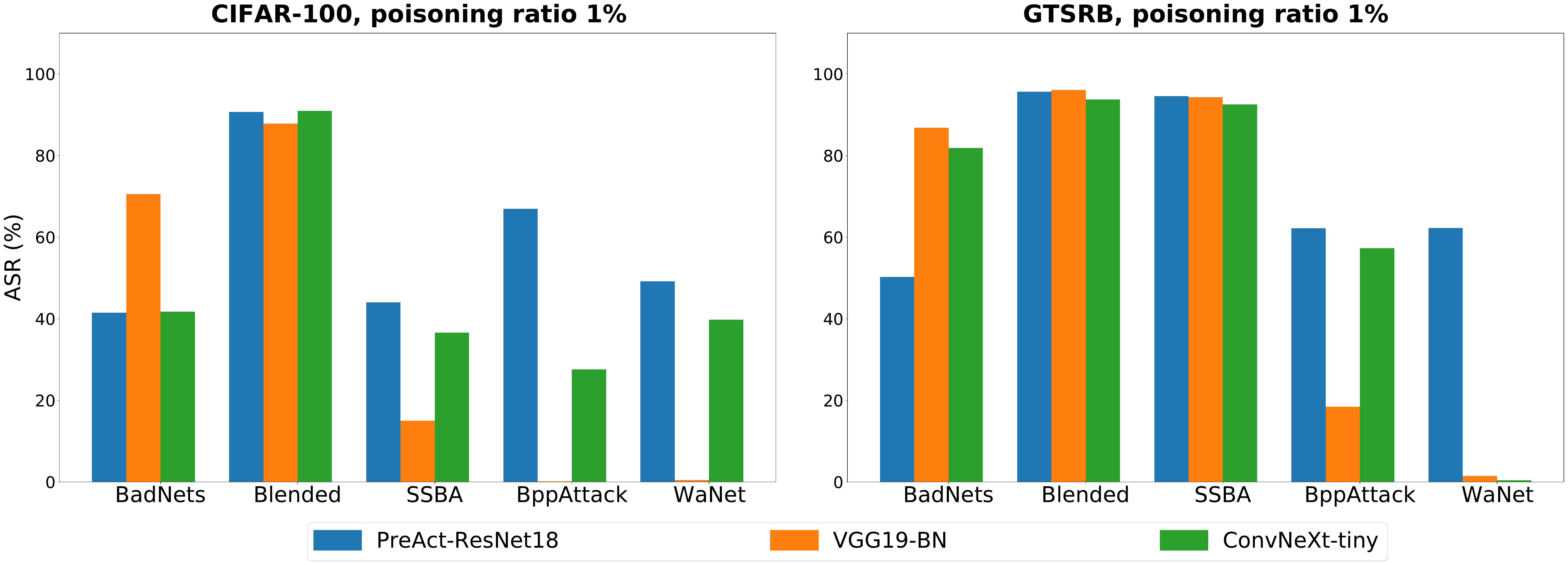}
\end{minipage}%
\centering
\vspace{-1em}
% \caption{Attack success rate (ASR) under different datasets, attacks, and model architectures.}
\caption{The influence of different model architectures measured by ASR.}
\label{fig:architecture}
\vspace{-1em}
\end{figure}

The choice of model architecture plays a pivotal role in the learning process. Particularly, ANP \cite{wu2021adversarial}, EP \cite{ep-bnp-2022}, and BNP \cite{ep-bnp-2022} methods necessitate harnessing information from the model's batch normalization layer. We consistently investigate the impact of diverse architectures on various attack types.
Our experimental setup encompasses five attacks, including two static attacks (\ie,  BadNets \cite{gu2019badnets} and Blended \cite{chen2017targeted}), and three dynamic attacks (\ie, SSBA \cite{ssba}, WaNet \cite{nguyen2021wanet}, and BppAttack \cite{wang2022bppattack}). The three models employed are PreAct-ResNet18, VGG19-BN, and ConvNeXt-tiny. In Figure~\ref{fig:architecture}, the ASR of these backdoor attacks across different models is illustrated. For static attacks, Blended exhibits similar ASR across various models, whereas BadNets demonstrates a robust attack effect, especially under VGG19-BN. Conversely, for dynamic attacks, VGG19-BN generally exhibits inferior performance compared to other networks with residual structures.

% 首先我们确定activation image,是指对于当前层的每一个神经元，找出最能激活该神经元的top-k images，之后我们计算poisoned sample占这些top images的比例，作为该神经元的poisoned ratio。最后，我们每层选出后门激活比例最高的神经元，poisoned image占的比例作为该层的activation ratio。

\begin{figure*}[htbp]
\centering
\begin{minipage}[t]{0.95\linewidth}
\centering
\includegraphics[width=\textwidth]{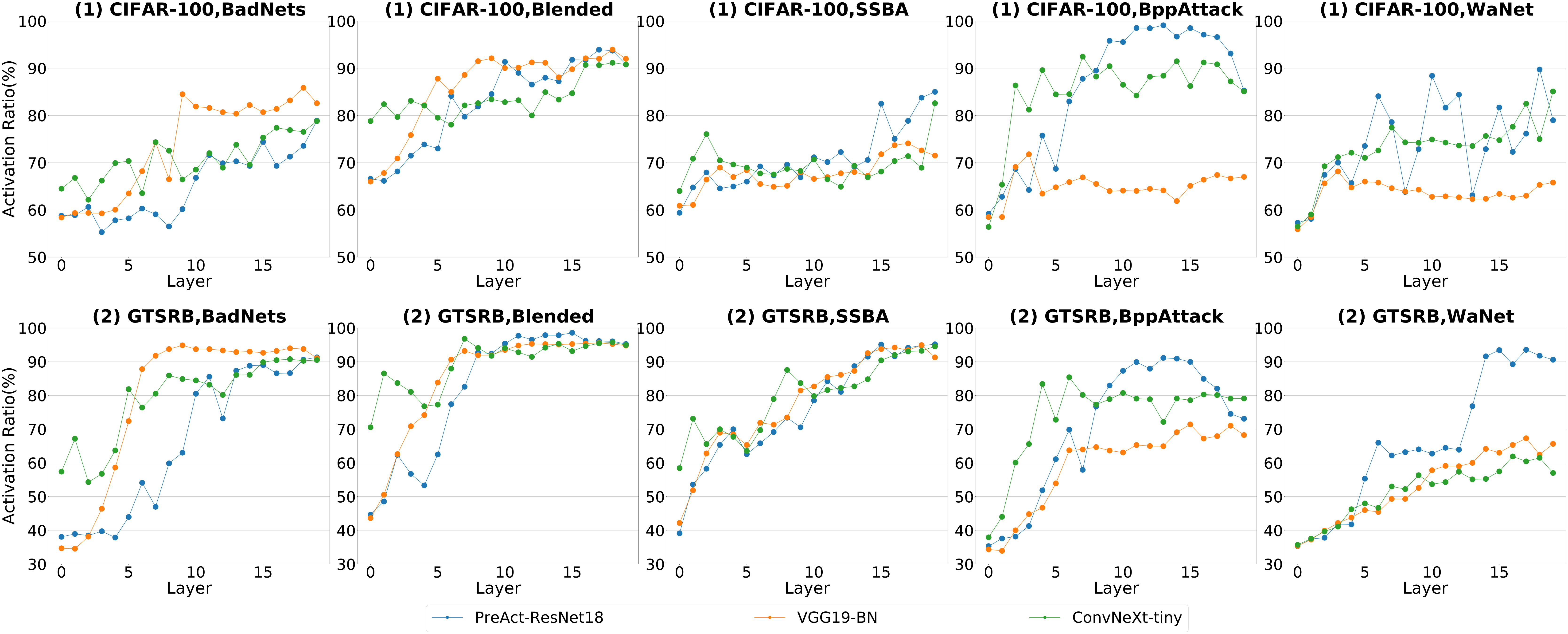}
\end{minipage}%
\centering
\vspace{-1em}
\caption{Backdoor activation ratio (BAR) under different datasets, attacks, and model architectures, with the poisoning ratio being 1$\%$ for all cases.}
\label{fig:architecture analysis}
\vspace{-1em}
\end{figure*}

To further delve into above observations, here we define a novel metric called \textit{backdoor activation ratio (BAR)}. It is calculated as follows. Firstly, given a set of images (50$\%$ clean and 50$\%$ poisoned) and one trained model, for each neuron, we record its activation values for all images, and pick $k$ images with the top-$k$ largest activation values. We calculate the ratio of poisoned images among these $k$ images for each neuron. Then, the largest ratio among all neurons for one layer is reported as the BAR score for that layer. BAR ranges from $0$ to $1$, and a higher score indicates stronger correlation between backdoor activation and one layer. The BAR scores under different datasets, attacks, and model architectures are presented in Figure~\ref{fig:architecture analysis}. 
It is very interesting to find that for each attack, the areas under the BAR curve of three model architectures are highly consistent with their ASR scores (see Figure \ref{fig:architecture}). It implies that the proposed BAR metric is a good tool to study the backdoor effect. Notably, the BAR curves of VGG19-BN are rather varied across different datasets and attacks, and sometimes the curve is very low, even in top layers (see the last two columns in Figure~\ref{fig:architecture analysis}). In contrast, the BAR curves of another two model architectures are much more stable. 
One critical difference between VGG19-BN and another two models is that the former doesn't adopt the residual connection. It implies us that the residual connection may amplify the backdoor activation in forward pass. 
Considering the complexity and diversity of model architectures, here we only present some preliminary observations and analyses, in order to inspire more researchers for deeper investigations about the influence of model architectures in backdoor learning.  

% We analyze the activation of neurons to uncover disparities in backdoor neuron activation across different architectures. As shown in Figure~\ref{fig:architecture analysis}, it displays the highest activation ratio of backdoor samples in each neural network layer. Our findings indicate a reduction in the activation ratio of dynamic backdoor attacks under VGG19-BN, particularly on the CIFAR-100 dataset. This phenomenon may be attributed to VGG19-BN's absence of a residual structure, resulting in excessively high dimensionality and scattered backdoor activation. Concurrently, we observed a higher activation ratio of BadNets under VGG19-BN, potentially due to VGG19-BN's traditional convolutional neural network design, which excels in extracting information from modules such as white blocks. This analysis underscores the distinctive characteristics of backdoor attacks in different modules, prompting further exploration of the influence of model structure on backdoor attack/defense algorithms.

%% file: sections/analysis/sharpness.tex
\paragraph{The sharpness of backdoored model}
\label{sec: Sharpness}
% 首先介绍一下sharp的作用
% 再说明我们使用的工具
% 再介绍一下实验设置
% 最后分析

\begin{figure*}[htbp]
% \centering
% \subfigure[Poisoned data]{
\centering
\begin{minipage}[t]{\linewidth}
\centering
\includegraphics[width=\textwidth, trim=2mm 2mm 2mm 2mm, clip]{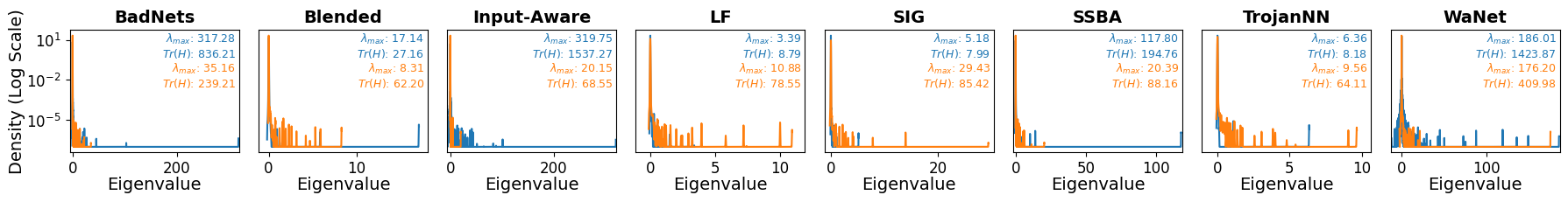}
\end{minipage}%
\label{fig:sharpness_attack(a)}
% }%
% \vspace{-3mm}
% \subfigure[Clean data]{
% \centering
% \begin{minipage}[t]{\linewidth}
% \centering
% %  \includegraphics[width=\textwidth, trim=left bottom right top, clip]{fig.png}
% \includegraphics[width=\textwidth, trim=2mm 2mm 2mm 2mm, clip]{figs/sharpness/Fig_attacks_Clean.pdf}
% \end{minipage}%
% \label{fig:sharpness_attack(b)}
% }%
% \vspace{-3mm}
% \subfigure[Mixed data]{
% \centering
% \begin{minipage}[t]{\linewidth}
% \centering
% \includegraphics[width=\textwidth, trim=2mm 2mm 2mm 2mm, clip]{figs/sharpness/Fig_attacks_Mixed.pdf}
% \end{minipage}%
% }%
\centering
\vspace{-1.5em}
\caption{
The eigenvalue density plots of the Hessian matrix, as well as its maximum eigenvalue ($\lambda_{max}$) and trace $Tr(H)$, are approximated based on  \textcolor{darkblue}{\textbf{poisoned samples}} and \textcolor{darkorange}{\textbf{clean samples}}, for backdoored models under different attacks on CIFAR-10 with PreAct-ResNet18, while the poisoning ratio is set to 10$\%$ for each attack.}
\label{fig:sharpness_attack}
\end{figure*}

\begin{figure*}[htbp]
\centering
% \subfigure[BadNets]{
% \begin{minipage}[t]{0.5\linewidth}
% \centering
% \includegraphics[width=\textwidth, trim=2mm 2mm 2mm 2mm, clip]{figs/sharpness1/Fig_defenses_badnet_Mixed.png}
% \end{minipage}%
% \label{fig:sharpness_defense_bad(a)}
% }%
% \subfigure[Blended]{
\begin{minipage}[t]{1\linewidth}
\centering
\includegraphics[width=\textwidth, trim=2mm 2mm 2mm 2mm, clip]{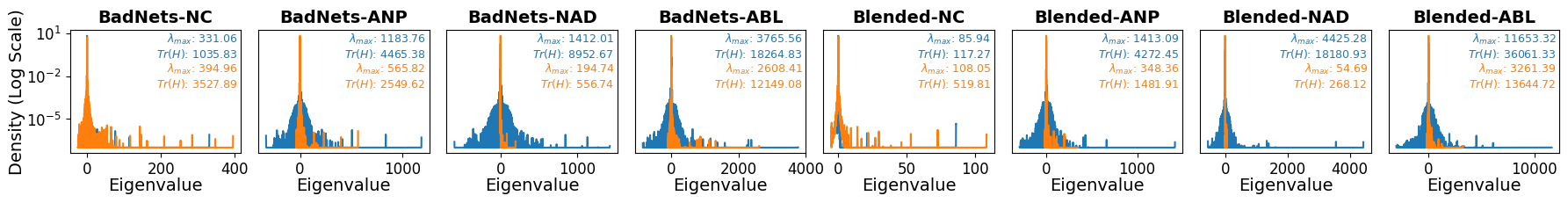}
\end{minipage}%
\label{fig:sharpness_defense_bad(b)}
% }%
\centering
\vspace{-1.5em}
\caption{The eigenvalue density plots of the Hessian matrix, as well as its maximum eigenvalue ($\lambda_{max}$) and trace $Tr(H)$, are approximated based on \textcolor{darkblue}{\textbf{poisoned samples}} and \textcolor{darkorange}{\textbf{clean samples}}, for backdoored models after different defenses against BadNets and Blended attack on CIFAR-10 with PreAct-ResNet18, while the poisoning ratio is set to 10$\%$ for each attack-defence pair.}
% \caption{Eigenvalue Density plots of Loss Hessian for \textcolor{darkblue}{\textbf{poisoned sample}}, and \textcolor{darkorange}{\textbf{clean sample}} on models after different defenses against BadNets and Blended attack on CIFAR-10 with PreAct-ResNet18. The maximum eigenvalue ($\lambda_{max}$) and the trace of the Hessian matrix ($Tr(H)$) for both types of sample are reported.}
\label{fig:sharpness_defense_bad}
\vspace{-1em}
\end{figure*}

% % Blended
% \begin{figure*}[htbp]
% \centering
% \subfigure[Blended, poisoned data]{
% \begin{minipage}[t]{0.5\linewidth}
% \centering
% \includegraphics[width=\textwidth, trim=2mm 2mm 2mm 2mm, clip]{figs/sharpness/Fig_defenses_blended_Poisoned.pdf}
% \end{minipage}%
% \label{fig:sharpness_defense_ble(a)}
% }%
% \subfigure[Blended, clean data]{
% \begin{minipage}[t]{0.5\linewidth}
% \centering
% \includegraphics[width=\textwidth, trim=2mm 2mm 2mm 2mm, clip]{figs/sharpness/Fig_defenses_blended_Clean.pdf}
% \end{minipage}%
% \label{fig:sharpness_defense_ble(b)}
% }%
% \centering
% \caption{Eigenvalue Density plots of Loss Hessian for poisoned data (a), and clean data (b) on models after different defenses against Blended attack on CIFAR-10 with PreAct-ResNet18. In each subfigure, the maximum eigenvalue ($\lambda_{max}$), trace of Hessian matrix ($Tr(H)$) are reported.}
% \label{fig:sharpness_defense_ble}
% \end{figure*}

The \textit{sharpness} (or \textit{flatness}) of the loss landscape is a widely observed characteristic to explore the model's mechanism or property, such as its relationship with model generalization \cite{hochreiter1997flat,keskar2016large,neyshabur2017exploring}. Here, we adopt it to study the backdoored models' mechanism. 
Specifically, to approximate the sharpness of one given model, we randomly select 1024 images from CIFAR-10, based on which we calculate the Hessian matrix $H$. Then, we compute the maximum eigenvalue $\lambda_{max}$, the trace $Tr(H)$, and plot the density of eigenvalues, using the Python package PyHessian\footnote{\url{https://github.com/amirgholami/PyHessian}}. Higher $\lambda_{max}$ and higher $Tr(H)$ tell that the model locates at a sharper point in the loss landscape. 
Note that the sharpness approximation depends on the adopted data, thus we conduct the approximation using poisoned data and clean data separately.

% In this section, we analyze backdoored model from a sharpness perspective. Many works suggest that generalization is correlated with the flatness of the loss landscape at the learned model, where a flatter minima usually implies a better generalization ability \cite{hochreiter1997flat,keskar2016large,neyshabur2017exploring}.
% Thus, it can be used to analyze how attacks and defenses work on the model from a sharpness perspective. In the following experiment in each group, we randomly select 1024 images and compute the maximum eigenvalue $\lambda_{max}$, the trace $Tr(H)$, and the eigenvalue density distribution of the Hessian matrix of the model with the given images using the PyHessian\footnote{\url{https://github.com/amirgholami/PyHessian}} package in Python. A higher $\lambda_{max}$ and higher $Tr(H)$ represent a sharper minimum. We study the influence of sharpness among attacks and defenses. Since the metric of sharpness is related to the input data, we compute the sharpness of the model using poisoned data and clean data separately.
% 对比每组的clean与poison可以看到，除了SSBA和input和badnet，其余的poison的都很小，这部分解释了trigger的泛化性极强的原因。SSBA和input是SSBA的trigger，因此可以理解他们和自然图像相比，语义特殊，处在landsca 比较sharp的区域。blended比较接近，最特殊的是badnet.一个合理的解释可能是badnet的激活通路极强，对应的网络lipschitz很大，所以landscape比较陡峭。
% 虽然mix是最接近网络学习的分布的，但mixed平坦度通常介于二者之间，这某种程度表明二者的学习是一种竞争关系
% 另外值得指出的是，在不同的组别之间，sharp程度和trigger的泛化性并不成正比，尽管sharp和网络泛化程度有关，但决定hessian大小的也和训练数据集有关系。

\textbf{Sharpness under different attacks.}
As observed in Figure \ref{fig:sharpness_attack}, most backdoored models locate at relatively flat points of the loss landscape for clean samples, while the situation is more complex for poisoned samples. 
For backdoors with sample-agnostic triggers, the poisoned landscape is even flatter than that the clean landscape (\eg, LF, SIG, and TrojanNN), which partly explains the high generalization of these triggers. 
However, for sample-specific triggers (\eg, Input-Aware, SSBA, and WaNet), the poisoned landscape can be very sharp. 
One exception is BadNets, which is also sample-agnostic, but its poisoned landscape is very sharp. 
One possible reason is that there is a high activation path in the network for BadNets, leading to large Lipschitz scores of some layers, thus the landscape becomes very sharp. 
Besides, the clean landscape of WaNet is also very sharp, with $\lambda_{max}$ being 176.2. It implies that backdoor could also affect the learning of clean samples.

\textbf{Sharpness under different defenses.}
Here, we study the impact of different defense methods on the sharpness of models. We study four different defenses against BadNets and Blended attacks and the results are shown in Figures \ref{fig:sharpness_defense_bad}. 
Compared to the plots in Figure \ref{fig:sharpness_attack}, it is clearly observed that both clean and poisoned landscapes become sharper, especially under ABL and ANP. 
For NC and NAD, the clean landscape is flat, implying the milder backdoor mitigation of tuning-based backdoor defenses. 
Besides, one notable issue is that measuring sharpness via Hessian matrix based on poisoned data makes no sense for defended models, because they are no longer local minimums for poisoned data (\ie, the first derivative is not close to zero). 

\textbf{In summary}, the sharpness is a highly discriminative metric for studying the difference among different backdoor attack and defense algorithms, while its accurate measure and relationship to model performance should be further explored. 

% We can find that for both clean data and poisoned data, the landscapes become sharper, especially for ABL and ANP. It is considerable since ABL removes backdoors by unlearning the model on suspicious poisoned data. For tuning-based methods like NC and NAD, the landscape for clean data is flatter, showing the milder backdoor mitigation of tuning-based backdoor defenses. Another important thing is that measuring sharpness for poisoned data on defense models does not make much sense because the network is no longer the minimum for poisoned data (the first derivative is not close to zero). Therefore, the Hessian matrix is of little significance to measure Sharpness.
% 这里使用的clean的数据为
% 从defense后模型的sharpness可以分析防御的机理，比如，ANP和ABL给网络带来了很大的扰动。而NC和NAD则相对平稳。poison的sharp非常大 其实也有yijiedao的作用。因为其在poison上已经不收敛了。
%%%%%%% attacks

%% file: sections/analysis/sensitivity.tex
% \paragraph{Analysis of sensitivity}
% \label{sec: analysis of individual_attack_method}

\begin{figure*}[htbp]
\centering
\includegraphics[width=\textwidth]{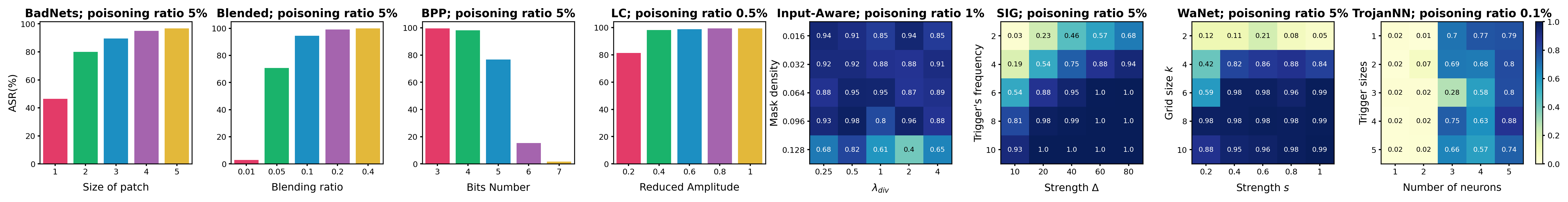}
\vspace{-1.5em}
\caption{The effects of attack methods under different hyper-parameters on CIFAR-10 with PreAct-ResNet18. For attacks where two hyper-parameters are adjusted, each axis represents a distinct hyper-parameter and ASR is marked in each box, the color of the box represents the level of ASR.}
\label{fig:sensiticity_attack}
\vspace{-1em}
\end{figure*}

\begin{figure*}[htbp]
\centering
\includegraphics[width=\textwidth]{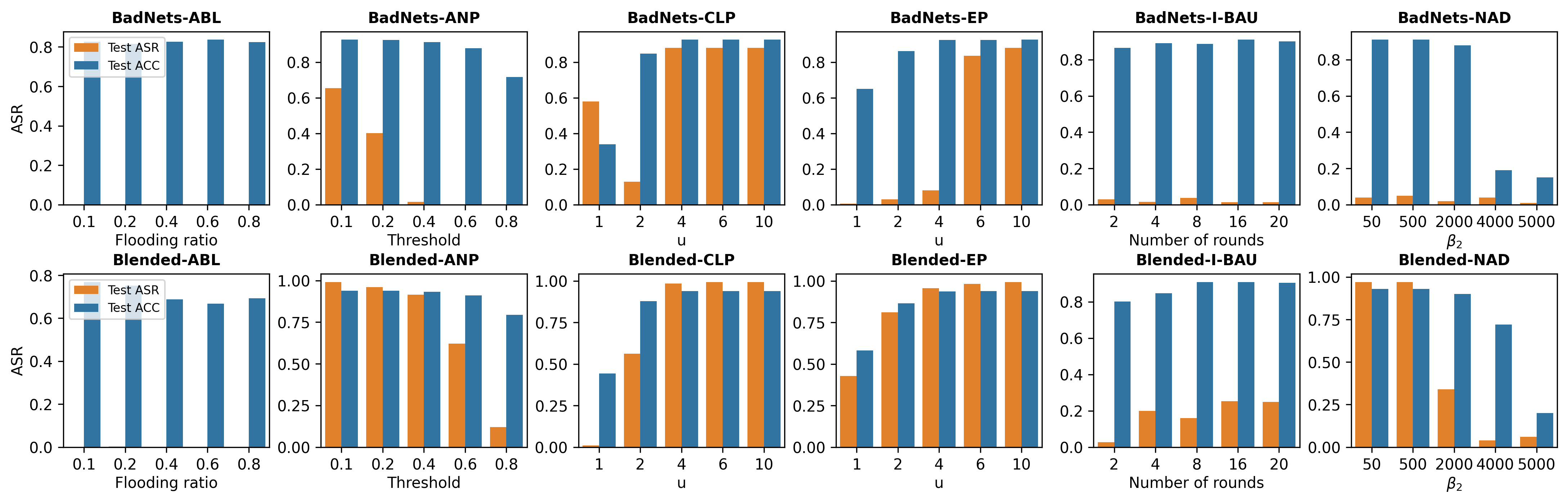}
\vspace{-2em}
\caption{The effects of defense methods under different hyper-parameters on CIFAR-10 with PreAct-ResNet18, while the poisoning ratio is set to 5$\%$ for both BadNets and Blended. Each row represents the performance of a different defense against the same attack and is indicated in the title.}
\label{fig: sensitivity_defense}
\vspace{-1em}
\end{figure*}

% \red{this subsection needs to be updated}

% In this section, we investigate the sensitivity of various attack and defense methods, taking into consideration different related hyper-parameters. Although a part of these hyper-parameters has been analysed, we analyse these parameters from a global perspective, such as the strength of the trigger, and the interaction of multiple parameters. 

\paragraph{Sensitivity analysis of each individual attack algorithm} 
% We experiment on the CIFAR-10 dataset with Preact-Resnet18, and adjust one or two critical hyper-parameters for each individual attack algorithm, with different poisoning ratios. 
For each attack algorithm, we adjust one or two critical hyper-parameters which are closely related to the trigger strength, to record the fluctuation of its attack performance, as shown in Figure~\ref{fig:sensiticity_attack}.
% However, as hyper-parameter changes do not affect their ASR under this poisoning ratio, we further analyze their attack effects at alternate poisoning ratios. 
% The results are shown in Figure~\ref{fig:sensiticity_attack}. 
Specifically, for the first four algorithms, we choose one hyper-parameter, such as the patch size of BadNets~\cite{gu2019badnets}, and the blending ratio of Blended~\cite{chen2017targeted}. 
For BPP~\cite{wang2022bppattack}, we manipulate the number of bits in the squeezed color palette within the range of 3 to 7. A lower number of bits implies a larger change in the original image, leading to an elevated ASR. 
For LC~\cite{turner2019labelconsistent}, we adjust its reduced amplitude between 0.1 and 1, which controls the trigger visibility. When the reduced amplitude falls below 0.2, the ASR decreases significantly. 
Their fluctuations demonstrate the strong impact of trigger strength on attack performance. 
For the remaining four algorithms, we choose two hyper-parameters, and thus we present an ASR matrix for sensitivity analysis. 
Although each hyper-parameter directly or indirectly affects the trigger strength, their impacts on ASR are different. For Input-Aware~\cite{nguyen2020input} and SIG~\cite{SIG}, the impacts of two hyper-parameters on ASR show an additive effect. In both cases, the ASR at one corner is higher and the ASR at the opposite corner is lower. 
For WaNet~\cite{nguyen2021wanet} and TrojanNN~\cite{liu2018trojaning}, there is one major hyper-parameter that directly affects the ASR. Only when this major hyper-parameter reaches a certain level, the ASR will be relatively high and another hyper-parameter may have an impact on ASR. The grid size of WaNet~\cite{nguyen2021wanet} and the number of trojaned neurons of TrojanNN~\cite{liu2018trojaning} serve as the major hyper-parameter, respectively.

\paragraph{Sensitivity analysis of each individual defense algorithm}
% We evaluate the performance of defense methods under different hyper-parameters on CIFAR-10 with Preact-Resnet18, aiming to investigate their sensitivity consistently. 
For each defense algorithm, we adjust one critical hyper-parameter to record the fluctuation of its defense performance, against BadNets~\cite{gu2019badnets} and Blended~\cite{chen2017targeted} attacks with a fixed poisoning ratio of 5\%. 
As shown in Figure~\ref{fig: sensitivity_defense}, some defense algorithms are sensitive to their hyper-parameters and can only achieve optimal results under specific settings. 
For tuning-based defense algorithms, such as ABL~\cite{li2021anti} and I-BAU~\cite{i-bau}, as they are all through unlearning or forgetting the identified or reversed poisoned samples, the selected hyper-parameter has minor influence on ASR and C-Acc. 
However, for NAD~\cite{nad-iclr-2020}, if it is too concentrated in the distillation task (\ie, a larger $\beta_2$), the C-Acc degradation is severe, but without a strong distillation, ASR cannot be reduced against Blended. 
For pruning-based defense algorithms, including ANP~\cite{wu2021adversarial}, CLP~\cite{clp-2022} and EP~\cite{ep-bnp-2022}, their pruning hyper-parameters have very strong impacts on ASR and C-Acc. 

\textbf{In summary}, above sensitivity analysis 
demonstrates that most attack/defense algorithms are sensitive to a few critical hyper-parameters.

%% file: sections/analysis/quick_learning_forget.tex
% \paragraph{The difference between clean and poisoned samples during the training process}
\paragraph{Learning speed of clean and poisoned samples}
\label{sec: learning difference on learning speed}

\begin{figure*}[htbp]
\centering
\includegraphics[width=\textwidth]{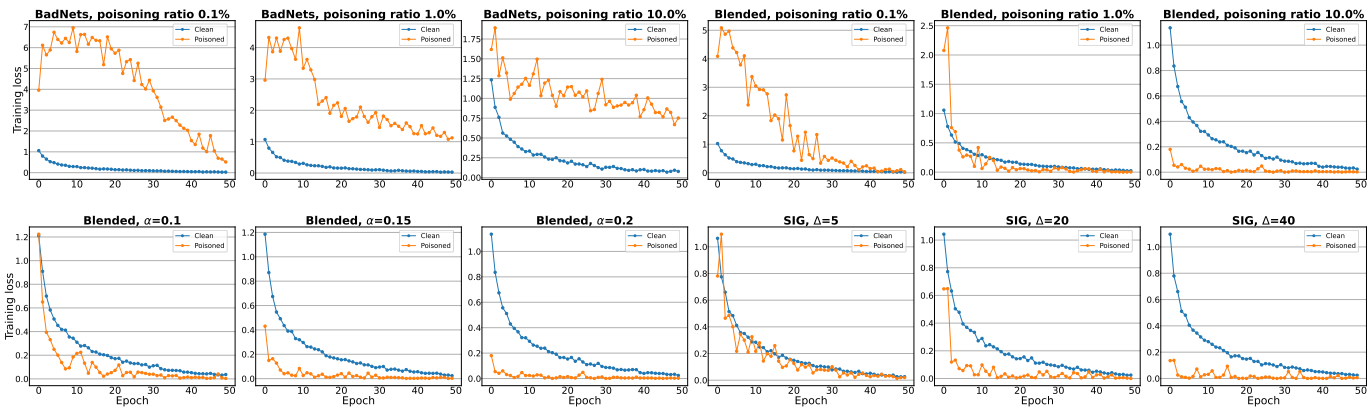}
\vspace{-1.7em}
\caption{Training loss curves of poisoned and clean samples under different backdoor attacks, with various poisoning ratios and trigger strengths. Note that $\alpha$ is set to 0.2 for Blended shown in the top row, while the poisoning ratio is set to $10.0\%$ for both Blended and SIG shown in the bottom row.}
\label{fig:quick_learn}
\vspace{-0.6em}
\end{figure*}

\begin{figure*}[htbp]
\centering
\includegraphics[width=\textwidth]{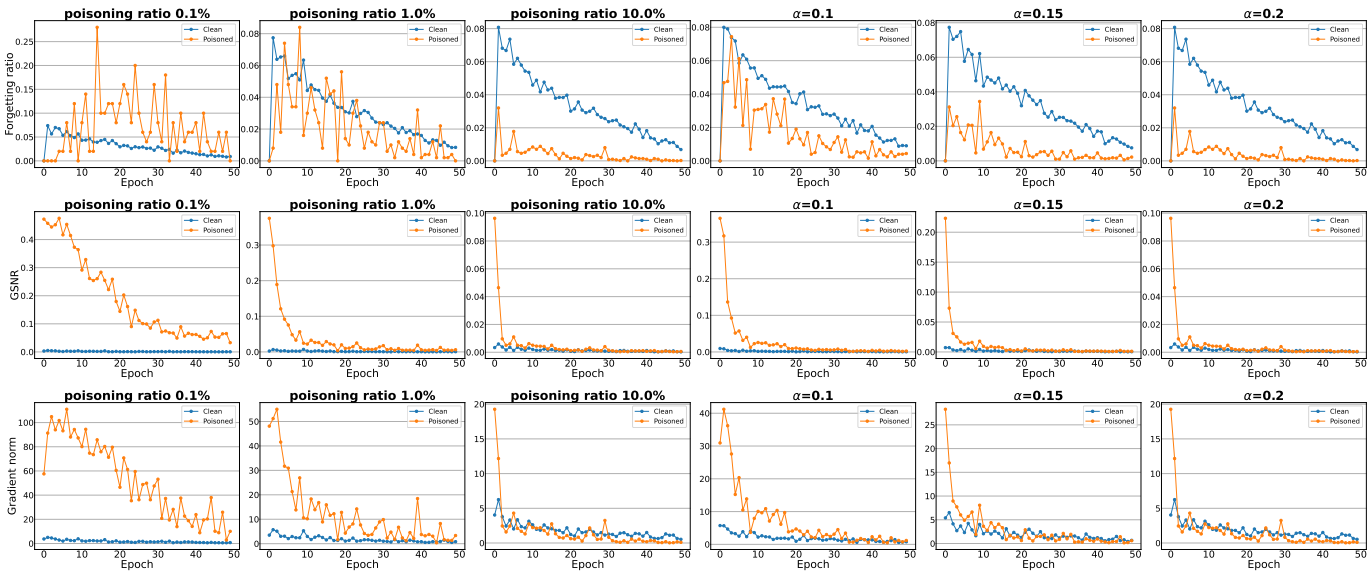}
\vspace{-1.5em}
\caption{Memory-related records during training of the evaluations on CIFAR-10 with PreAct-ResNet18, under the Blended attack, with various poisoning ratios and trigger strengths (reflected by $\alpha$). Note that  $\alpha$ is set to $0.2$ in the first three column, while the poisoning ratio is set to $10.0\%$ in the last three columns.}
\label{fig:quick_learn_ana}
\vspace{-1em}
\end{figure*}

It has been widely observed in backdoor learning that the poisoned samples will be learned more quickly than the clean samples, called \textit{the quick learning phenomenon}. 
To further verify this phenomenon and reveal the underlying mechanism, we evaluate on CIFAR-10 with PreAct-ResNet18 under three backdoor attacks, with various poisoning ratios and trigger strengths (reflected by the hyper-parameter $\alpha$ in Blended and $\delta$ in SIG). 
As shown in Figure~\ref{fig:quick_learn}, the quick learning phenomenon is very clear in most cases that the loss reduction rate of poisoned samples is much higher than that of clean samples, and the reduction rate is positively related with the poisoning ratio and trigger strengths. However, there are exceptions for BadNets, where the poisoned loss reduction is even slower than the clean loss reduction, though the backdoor is finally well learned with high ASRs on poisoned testing samples.  
We believe the possible reason is that the patch trigger of BadNets is too small to be resistant to the sample augmentation during the training process, causing its slow learning speed.  
\textbf{In short}, the quick learning phenomenon reveals that the DNN model prefers to firstly learn the simple mapping from the trigger to the target class, then learn the more complex mappings from the objects to their semantic classes.

\paragraph{Memorization}
\label{sec: learning difference on memorization}

In the following, we provide a detailed analysis from the perspective of memorization. Specifically, in each epoch during the training process, we record the following memory-related information: 
\begin{itemize}
    \item \textit{Forgetting ratio}: a forgetting event~\cite{forgetting} is recorded when one training sample is correctly predicted to its given label in the current epoch, but incorrectly predicted in the next epoch. We calculate the forgetting ratios of all poisoned and all clean samples, respectively. 
    \item \textit{Gradient signal to noise ratio (GSNR)} \cite{liu2020understanding}: 
    The GSNR of one parameter denotes the ratio between its gradient’s squared mean and variance, over the data distribution (or a set of samples). It measures the concentration of a parameter’s gradients among different training samples, and large GSNR values indicate that the update directions for that parameter from most training samples are more similar. We record the GSNRs averaged over all model parameters, among poisoned and clean training samples, respectively;
    \item \textit{Gradient norms averaged over all model parameters}: the average gradient norms on all training samples, clean training samples, and poisoned training samples are recorded, respectively. 
    % \item Pairwise cosine similarities between average gradients on total training samples, clean training samples, and poisoned training samples.
\end{itemize}

% 从图像中clean sample和poisoned sample的memory-related information的对比可以看出，投毒样本的梯度相较于干净样本都非常大，而且GSNR比干净样本大的时间更长，同时，投毒样本一旦被记住就很难被模型遗忘。这说明投毒样本具有统一的特征，使得模型更容易也更加关注于投毒样本的学习。同时从不同列中poisoned sample的表现可以看出，投毒样本越多，投毒的trigger越强，反倒是梯度norm和GSNR对于poisoned training sample越小。我们认为这是由于模型更加容易学习到poisoned sample，所以模型对于poisoned training sample的关注越小，但是还是远大于clean training sample。

Our evaluations are conducted on CIFAR-10 with PreAct-ResNet18, under the Blended attack, with various poisoning ratios and trigger strengths (reflected by $\alpha$). The memory-related records are shown in Figure \ref{fig:quick_learn_ana}.  
Firstly, according to the sub-plots in the top row, we observe that the forgetting ratio of poisoned samples decreases along with both the poisoning ratio and the trigger strength (\ie, $\alpha$), and when these two factors are sufficiently large, the poisoned forgetting ratio will be lower than the clean forgetting ratio. As the lower forgetting ratio implies that more samples are stably fitted, this observation is fully consistent with the above observations about the training losses (see the subplots of Blended in Figure \ref{fig:quick_learn}). 
Then, according to the sub-plots in the middle row, we observe that the poisoned GSNR is much larger than the clean GSNR at all cases of poisoning ratio and $\alpha$, especially at early epochs. We believe the reason is that all poisoned samples are guided to the same target class, while all clean samples are guided to different classes, thus the update directions of poisoned samples should be more concentrated than those of clean samples. 
Last, according to the sub-plots in the bottom row, we observe that the gradient norms of poisoned samples are much larger than those of clean samples at all cases of poisoning ratio and $\alpha$, especially at early epochs. One reasonable reason is that since poisoned samples are picked from different source classes, their appearances are very dispersed. When they are guided towards the same destination, the gradient norms must be very large. 
Moreover, when the poisoning ratio is low (\eg, $0.1\%$), although both GSNR and gradient norm of the poisoned samples are large, the overall update strength from a limited poisoned samples is still weak, such that the poisoned samples cannot dominate the model updates. All these records about GSNR and gradient norms perfectly explain the above observations about the quick learning phenomenon and forgetting ratio. We believe that these analyses could shed light on the backdoor learning mechanism, to facilitate us to develop more effective backdoor learning algorithms or build backdoor related theories in future.

%% file: sections/conclusion.tex
\section{Conclusions and future plans}
\label{sec: conclusion}

\textbf{Conclusions}. We have established a comprehensive benchmark of backdoor learning, called \textbf{BackdoorBench}, encompassing an extensible modular-based codebase with the implementations of 20 backdoor attack and 32 backdoor defense algorithms, and 11,492 pairs of attack-against-defense evaluations, as well as extensive and in-depth analyses supported by 18 analytical tools. 
BackdoorBench has not only presented a clear overview of the current progress of backdoor learning, but also provided an user-friendly toolbox and resource to facilitate researchers to develop and verify more advanced backdoor learning algorithms, theories or applications. 
However, we believe its most significant contribution to the research community is that it has shed light on the intrinsic characteristics and mechanisms of backdoor learning based on extensive analyses, which may appeal and inspire more researchers to together promote this topic's development.  
 
% \textbf{Conclusions}. In conclusion, our work has established BackdoorBench, a comprehensive benchmark for backdoor learning, addressing the existing challenges in the field. By providing an extensible modular-based codebase encompassing 16 advanced backdoor attacks and 21 defense algorithms, along with 11,492 conducted evaluations and thorough analyses from 10 aspects, we aim to contribute significantly to the backdoor community.
% Our benchmark serves as a crucial resource, offering a clear overview of the current state of backdoor learning. It empowers researchers by enabling quick comparisons with existing methods while developing new approaches. Furthermore, the comprehensive evaluations conducted and the insights derived from the analyses inspire the exploration of new research problems in the realm of backdoor learning.

\textbf{Future plans}. 
BackdoorBench will be continuously developed and maintained as a long-term project, supported by a strong and stable research team. 
Our major extension plan is elaborated from two dimensions. \textit{One dimension} is application scenario. Our current algorithms and evaluations mainly focus on computer vision, and we plan to cover a wider range of application scenarios, especially the security or safety critical ones, such as natural language processing, speech recognition, intelligent robotics, \etc.  
\textit{Another important dimension} is model. currently we mainly focus on discriminative models (\ie, classification), and we plan to extend to generative models (\eg, large language models, large vision models, multimodal large language models, diffusion models), and foundation models (\ie, CLIP model). These models become very popular in recent years as they can serve as the strong basis of several down-streaming tasks or applications. Thus, investigating their security issues has very important research and practical values.